\documentclass[acmlarge,screen,prologue,table,xcdraw]{acmart}
\usepackage{xcolor}
\usepackage{verbatim}
\usepackage{xspace}
\usepackage{algorithm}
\usepackage{units}
\usepackage{graphicx}
\usepackage{url}
\usepackage{array}
\usepackage{multirow}
\usepackage{caption}
\usepackage{subfig}
\usepackage{algpseudocode}
\usepackage{setspace}
\usepackage{bbm}
\usepackage{hyperref}
\usepackage{makecell}

\usepackage{booktabs}
\usepackage[normalem]{ulem}
\usepackage{aecompl}
\usepackage{dblfloatfix}
\usepackage{bm}
\usepackage{bbm}
\usepackage{dirtytalk}
\usepackage{nomencl}
\makenomenclature
\usepackage{subfig}
\usepackage{float}

\specialcomment{Notes}{%
	\noindent$\bf\Rightarrow\Rightarrow$%
}{%
	$\Leftarrow\Leftarrow$%
}
\providecommand{\Note}[1]{%
	\noindent{\bf$\Rightarrow\Rightarrow$#1$\Leftarrow\Leftarrow$}%
}
\newcommand{\Donothing}{\ignorespaces\ifhmode\unskip\fi}
\newcommand{\HideNotes}{%
	\excludecomment{Notes}%
	\renewcommand{\Note}[1]{\Donothing}%
}
\HideNotes
\usepackage{xcolor}

\usepackage{amsmath}
\DeclareMathOperator*{\argmax}{argmax}

\usepackage{amssymb}
\usepackage{amsfonts}    
\usepackage{nicefrac}      
\usepackage{microtype}      
\setcopyright{acmcopyright}
\acmJournal{IMWUT}
\acmYear{2022}\acmVolume{6}\acmNumber{1}\acmArticle{38}\acmMonth{3}
\acmDOI{10.1145/3517250}
\begin{document}
\title{Do Smart Glasses Dream of Sentimental Visions? Deep \textit{Emotionship} Analysis for Eyewear Devices}

\author{Yingying Zhao}
\authornote{Equal contribution}
\email{yingyingzhao@fudan.edu.cn}
\affiliation{%
  \institution{School of Computer Science, Fudan University}
  \city{Shanghai}
  \country{China}
  \postcode{200438}
}
\affiliation{%
  \institution{Shanghai Key Laboratory of Data Science, Fudan University}
  \city{Shanghai}
  \country{China}
  \postcode{200438}
}

\author{Yuhu Chang}
\authornotemark[1]
\email{yhchang14@fudan.edu.cn}
\affiliation{%
  \institution{School of Computer Science, Fudan University}
  \city{Shanghai}
  \country{China}
  \postcode{200438}
}
\affiliation{%
  \institution{Shanghai Key Laboratory of Data Science, Fudan University}
  \city{Shanghai}
  \country{China}
  \postcode{200438}
}

\author{Yutian Lu}
\email{20210240098@fudan.edu.cn}
\affiliation{%
  \institution{School of Computer Science, Fudan University}
  \city{Shanghai}
  \country{China}
  \postcode{200438}
}
\affiliation{%
  \institution{Shanghai Key Laboratory of Data Science, Fudan University}
  \city{Shanghai}
  \country{China}
  \postcode{200438}
}

\author{Yujiang Wang}
\authornote{Corresponding author}
\email{yujiang.wang14@imperial.ac.uk}
\affiliation{%
  \institution{Department of Computing, Imperial College London}
  \city{London}
  \country{United Kingdom}
}

\author{Mingzhi Dong}
\email{mingzhidong@gmail.com}
\affiliation{%
  \institution{School of Computer Science, Fudan University}
  \city{Shanghai}
  \country{China}
  \postcode{200438}
}
\affiliation{%
  \institution{Shanghai Key Laboratory of Data Science, Fudan University}
  \city{Shanghai}
  \country{China}
  \postcode{200438}
}

\author{Qin Lv}
\email{qin.lv@colorado.edu}
\affiliation{%
  \institution{University of Colorado Boulder}
  \city{Boulder}
  \state{Colorado}
  \country{United States}
}

\author{Robert P. Dick}
\email{dickrp@umich.edu}
\affiliation{%
  \institution{University of Michigan}
  \city{Ann Arbor}
  \state{Michigan}
  \country{United States}
}

\author{Fan Yang}
\email{yangfan@fudan.edu.cn}
\affiliation{%
  \institution{School of Microelectronics, Fudan University}
  \city{Shanghai}
  \country{China}
  \postcode{201203}
}

\author{Tun Lu}
\email{lutun@fudan.edu.cn}
\affiliation{%
  \institution{School of Computer Science, Fudan University}
  \city{Shanghai}
  \country{China}
  \postcode{200438}
}
\affiliation{%
  \institution{Shanghai Key Laboratory of Data Science, Fudan University}
  \city{Shanghai}
  \country{China}
  \postcode{200438}
}

\author{Ning Gu}
\email{ninggu@fudan.edu.cn}
\affiliation{%
  \institution{School of Computer Science, Fudan University}
  \city{Shanghai}
  \country{China}
  \postcode{200438}
}
\affiliation{%
  \institution{Shanghai Key Laboratory of Data Science, Fudan University}
  \city{Shanghai}
  \country{China}
  \postcode{200438}
}

\author{Li Shang}
\email{lishang@fudan.edu.cn}
\affiliation{%
  \institution{School of Computer Science, Fudan University}
  \city{Shanghai}
  \country{China}
  \postcode{200438}
}
\affiliation{%
  \institution{Shanghai Key Laboratory of Data Science, Fudan University}
  \city{Shanghai}
  \country{China}
  \postcode{200438}
}

\renewcommand{\shortauthors}{Zhao et al.}

\begin{abstract}
Emotion recognition in smart eyewear devices is highly valuable but challenging. One key limitation of previous works is that the expression-related information like facial or eye images is considered as the only emotional evidence. However, emotional status is not isolated; it is tightly associated with people's visual perceptions, especially those sentimental ones. However, little work has examined such associations to better illustrate the cause of different emotions.
In this paper, we study the \textit{emotionship} analysis problem in eyewear systems, an ambitious task that requires not only classifying the user's emotions but also semantically understanding the potential cause of such emotions. To this end, we devise \textit{EMOShip}, a deep-learning-based eyewear system that can automatically detect the wearer's emotional status and simultaneously analyze its associations with semantic-level visual perceptions. 
Experimental studies with 20 participants demonstrate that, thanks to the \textit{emotionship} awareness, \textit{EMOShip} not only achieves superior emotion recognition accuracy over existing methods (80.2\% vs. 69.4\%), but also provides a valuable understanding of the cause of emotions. Pilot studies with \textcolor{black}{20} participants further motivate the potential use of \textit{EMOShip} to empower emotion-aware applications, such as emotionship self-reflection and emotionship life-logging. 
\end{abstract}
%
%

\ccsdesc[500]{Human-centered computing~Mobile devices}
\keywords{Smart Eyewear System, Emotionship, Emotion Recognition, Sentiment Analysis, Image Captioning, Visual Question Answering}
\maketitle

\section{Introduction}
\label{sctn::intro}
Research in social and psychology science indicates that our emotional state can considerably affect different aspects of our daily life, including our thoughts and behaviors~\cite{costa2016emotioncheck}, decision making~\cite{ruensuk2020you}, cognitive focuses~\cite{10.1145/3264913}, performance on assessments~\cite{pekrun2011measuring}, physical health~\cite{di2018emotion}, and mental well-beings~\cite{tugade2004resilient}. Given the significant impacts of emotions, emotion recognition is one of the most crucial research topics in affective computing~\cite{pantic2003toward}, and it can be applied to a wide range of human-computer interaction (HCI) scenarios to improve user experience. Intelligent eyewear systems are especially well suited to carry out and benefit from emotion recognition. 

A common goal of smart eyewear devices is to deliver intelligent services with personalized experiences. This requires understanding the users, especially their affective status. 
As indicated by previous studies~\cite{costa2016emotioncheck, cowie2001emotion, aziz2008embodied, dalgleish2004emotional, zhao2016emotion}, the ability to recognize emotion can greatly enhance user experience in various HCI scenarios. More importantly, an emotion-sensitive wearable front-end would enable a variety of personalized back-end applications, such as emotional self-reflection \cite{10.1145/3264913,ghosh2020towards}, 
emotional life-logging \cite{blum2006insense}, emotional retrieving and classification \cite{yang2018retrieving}, and mood tracking \cite{torkamaan2020mobile}.

Recognizing emotions using smart eyewear devices is challenging. The majority of state-of-the-art emotion recognition techniques~\cite{ebrahimi2015recurrent, kim2015hierarchical, levi2015emotion, li2017reliable, li2020deep} use deep learning models to classify expressions from full facial images. However, it is typically difficult to capture the entire face using sensors that can economically be integrated into current eyewear devices. 
This mismatch between economical sensors and analysis techniques hinders the practical application of existing emotion recognition methods in eyewear.

To address this challenging problem, previous works \cite{scheirer1999expression, fukumoto2013smile, masai2015affectivewear} adopted engineering-based approaches to extract hand-crafted features from eye regions instead of the whole facial images to compute the affective status.
With the embedding of eye-tracking cameras in commercial eyewear devices, recent eyewear systems developed convolutional neural networks (CNN) to extract deep affective features from eye-camera-captured images (typically eye regions) for head-mounted virtual reality (VR) glasses \cite{hickson2019eyemotion} and smart glasses \cite{wu2020emo}. 
Besides the limited recognition accuracy, those prior works predict human emotions based on the expression information of eyes solely and exclusively, ignoring the subtle yet crucial associations between people's emotional status and visual perceptions. In fact, the hints to the user's emotional state can be discovered in both expressions and visual experiences. Learning additional sentimental clues in the latter will inevitably benefit emotion recognition from the former. 

As shown in studies of behaviors and neuroscience \cite{ohman2001emotion, compton2003interface, okon2014neural}, the sentimental content in the scene is generally prioritized by people's visual attention over those emotionally neutral ones. These emotional-arousing contents are also known as \textit{emotional stimuli}. For example, viewing a child playing with parents can lead to joyfulness, while we will feel sad if we perceive a crying woman who just lost her husband. In other words, emotion is not an isolated property; instead, it is tightly connected with the emotional stimuli of our visual attention. The arising of our emotions can be closely associated with the varying sentimental visions of our views, especially for eyewear devices with rapidly altering scenes.  

Based on such observations, we study the \textit{emotionship} 
analysis problem in eyewear devices. 
The term \textit{emotionship} conceptualizes the association of emotional status with the relevant hints in expression and visual attention. Through \textit{emotionship} analysis, we aim to recognize emotions with better accuracy and understand the semantic cause \footnote{\textcolor{black}{In this paper, the phrase \say{understanding the cause} refers to the understanding of the momentary associations between emotional states and the scene images in eyewear devices. It should be distinguished with the understanding of the emotions' causality \cite{coegnarts2016perceiving} which is a different task.}} 
for such emotions through a quantitative measurement of the emotional contributions from vision attentions. 
In this paper, we adopt the widely accepted emotion categorization system that classifies emotions into six basic categories \cite{ekman1992argument} plus the extra neutrality following \cite{wu2020emo} to define the status of emotions. It is important to note that a semantic-level understanding of visual experiences is necessary, since a certain attention region may consist of multiple objects and therefore can be ambiguous to establish the associations. In other words, we need to capture the semantic attributes of the visual perceptions. 
Compared with traditional emotion recognition techniques, the proposed \textit{emotionship} analysis is arguably more ambitious and more difficult, as additional challenges arise from the semantic analysis of the human visual perception, its association with the emotional status, etc. 
However, a successful \textit{emotionship} analysis framework will clearly lead to a truly personalized eyewear system that is capable of performing unseen and valuable emotion-aware downstream tasks.

\begin{figure}[t!]
	\includegraphics[width=1 \textwidth]{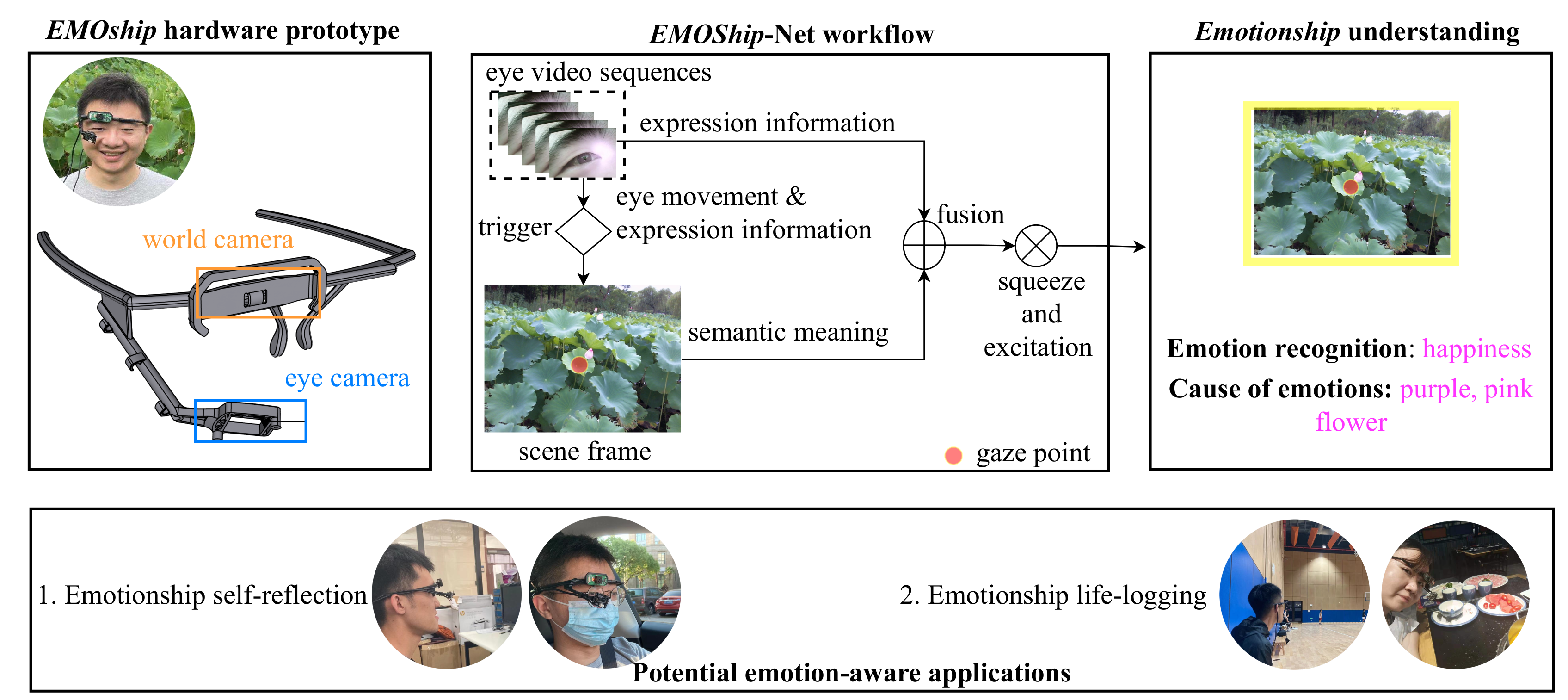}
	\caption[]{The proposed \textit{EMOShip} smart eyewear system.}
	\label{fig::framework}
\end{figure}

In this work, we present such an \textit{emotionship}-aware eyewear system for the first time. As shown in Fig. \ref{fig::framework}, our eyewear system, called \textit{EMOShip},
is equipped with cutting-edge deep learning techniques and is capable of recognizing the semantic attributes of the visual attentive regions, the expression-related information in eye images, and the emotional states based on the associations from both pieces of evidence. 
At the heart of \textit{EMOShip} is \textit{EMOShip}-Net, a deep neural network that is designed to address the new challenges in \textit{emotionship} analysis. To extract the semantic attributes of visual perceptions, we combine gaze points from eye-tracking~\cite{kassner2014pupil} with the \textcolor{black}{visual features} model VinVL ~\cite{zhang2021vinvl} \textcolor{black}{plus a vision-language (VL) fusion model OSCAR+ \cite{zhang2021vinvl}}. The sentimental clues in visual perceptions are synthesized with the expression-related information in eye images to predict emotion status more accurately and robustly. Visual perception's contributions to emotional states, which is subtle and challenging to measure, is quantitatively evaluated by a Squeeze-and-Excitation (SE) network \cite{hu2018squeeze} that fuses the scene and eye features. 
To evaluate the in-lab performance of \textit{EMOShip}, we collect and construct a new dataset, named EMO-Film. 
With the availability of visual perceptions' semantic attributes, the emotional states and the emotional impacts of visual attentions, our smart glasses system \textit{EMOShip} outperforms baseline methods on EMO-Film dataset in terms of emotion recognition accuracy, and more importantly, \textit{EMOShip} provides a semantic understanding of the potential cause of such emotions.
In-field pilot studies have been conducted to illustrate the effectiveness and superiority of this \textit{emotionship}-aware eyewear systems, and demonstrate its potential applications to a number of \textit{emotionship}-relevant tasks such as emotionship self-reflection and emotionship life-logging.

In summary, this paper makes the following contributions.

\begin{enumerate}

\item{\textcolor{black}{This work designs a smart eyewear \textit{EMOShip} to measure the relationship between the semantic attributes of visual attention and the emotional status of the wearer, and using this learned relationship increases the accuracy of emotional state recognition.}}

\item{\textcolor{black}{The proposed \textit{EMOShip} is equipped with a deep neural network \textit{EMOShip}-Net, which is designed to extract expression-related affective features and sentimental clues in visual attention. Most importantly, \textit{EMOShip}-Net fuses them to achieve more accurate emotion recognition and quantify their \textit{emotionship} associations.} }

\item{On the self-collected EMO-Film dataset, \textit{EMOShip} achieves approximately 10.8\% higher emotion recognition accuracy than the baseline methods, and demonstrates its potential capability of providing valuable sentimental clues for \textit{emotionship} understanding.
}

\item{We perform in-field pilot studies on two inspiring down-stream applications -- emotionship self-reflection and emotionship life-logging, to illustrate the potential use of \textit{EMOShip}. With \textcolor{black}{three}-week studies on \textcolor{black}{20} participants, we have shown that \textit{EMOShip} has achieved \textcolor{black}{82.8\%} precision in terms of emotional moment capturing. The questionnaire survey shows that \textcolor{black}{16 out  of 20} users embrace the idea of emotionship self-reflection and admit its benefits, while \textcolor{black}{15} out of \textcolor{black}{20} users give positive feedback on emotion life-logging applications.
}
\end{enumerate}

The rest of the paper is organized as follows. Section \ref{sctn::related} surveys related works. Section \ref{sctn::system} presents the proposed \textit{EMOShip} algorithm design and system implementation. Section \ref{sctn::exp} presents the experimental results. Section \ref{sctn::app} demonstrates the in-field pilot studies. We conclude the work in Section \ref{sctn::cnclusn}. 

\section{Related Work}
\label{sctn::related}
This section surveys related work in the field of (1) emotion recognition, (2) image sentiment analysis, (3) vision-language models, and (4) gating mechanisms. We also highlight the key contributions of \textit{EMOShip} compared to prior work.

\subsection{Emotion Recognition}
Ekman proposed a well-known and widely-adopted emotion categorization system that divided emotions into six basic categories: happiness, sadness, fear, anger, disgust, and surprise~\cite{ekman1992argument}. And the seventh emotion is neutrality~\cite{wu2020emo}, which represents the absence of emotion. 
The seven basic emotions are widely accepted~\cite{aifanti2010mug,wu2020emo}, and this work also adopts this emotion categorization system. 

Most recent works involve deep models to classify the seven basic emotions from the whole facial images \cite{ebrahimi2015recurrent, kim2015hierarchical, levi2015emotion, li2017reliable, li2020deep}, as facial expressions are one of the most common channels for humans to express feelings \cite{garcia2017emotion}.
When it comes to smart eyewear devices, however, the whole facial images are not easy to obtain, and therefore alternatives should be established.
Eye region image has been shown to contain sufficient expression-related information \cite{wu2020emo}, and since it can be easily fetched through matured eye cameras, eye images and eye analysis techniques have become a promising choice for emotion recognition in eyewear systems.
Tarnowski et al. proposed to utilize eye-tracking information, mainly regarding the eye movements and pupil diameters, for emotion recognition \cite{tarnowski2020eye}. 
Aracena et al. presented an emotion recognition method based only on pupil size and gaze position \cite{aracena2015neural}. Later, Wu et al. proposed a deep-learning-based network to extract emotional features from the single-eye-area images, and classified them into seven basic emotions~\cite{wu2020emo}. 

In this work, we take one further step from the traditional emotion recognition and study the \textit{emotionship} analysis, a task that not only requires to learn the emotional states but also should consider the impacts of sentimental visual perceptions.

\subsection{Image Sentiment Analysis}
Different from emotion recognition based on facial expressions, visual sentiment analysis aims to predict the intended emotions from images. This work mainly investigates visual sentiment analysis based on categorical approaches that divide the intended emotions from images into six categories~\cite{she2019wscnet}, which is usually consistent with the emotion categorization system~\cite{ekman1992argument}.

Early sentiment prediction used hand-crafted features to recognize intended emotions. Those features included color variance, composition, and image semantics \cite{machajdik2010affective}, etc. Recently, with the advancing of deep Convolutional Neural Networks (deep CNNs), numerous deep-learning-based sentiment prediction approaches have been proposed to extract deep features for more effective sentiment prediction~\cite{yang2018weakly,she2019wscnet}. Campos et al. conducted extensive experiments and compared the performance of several fine-tuned CNNs for visual sentiment prediction \cite{campos2017pixels}. Zhu et al. proposed a unified CNN-RNN model to predict image emotions based on both low-level and high-level features by considering the dependencies of the features~\cite{zhu2017dependency}. Rao et al. classified image emotions based on a proposed multi-level deep network that combined the local emotional information from emotional regions with global information from the whole image~\cite{rao2019multi}. Yang et al. proposed a weakly supervised coupled convolutional network to provide effective emotion recognition by utilizing the local information of images~\cite{yang2018weakly}. Later, they extended the proposed weakly supervised detection framework through a more detailed analysis for visual sentiment prediction~\cite{she2019wscnet}. 

We acquire from those works the idea of image sentimental analysis and extract the sentimental features in scene images through a Vision-Language (VL) model. 

\subsection{Vision-Language Models} 
Vision-Language (VL) models are a relatively new field in computer vision, and they are designed for the VL tasks \cite{lu2019vilbert,tan2019lxmert, chen2019uniter,li2020unicoder,zhou2020unified}. VL models usually consist of two stages: 1. An object detection model is involved to predict the Region of Interests (RoIs) of each object and also to extract the feature embedding for each RoI, 2. a cross-modal fusion model to generate short descriptions of each RoI's semantic attributes. Therefore, a successful VL model will generate all RoIs of a scene image, the feature embedding for each RoI, and also the semantic attributes per RoI. 
VinVL model \cite{zhang2021vinvl} improves the performance of the vision module to extract visual presentations at higher qualities, and employs OSCAR \cite{li2020oscar} which is based on transformer \cite{vaswani2017attention} to perform the cross-modal semantic attributes predictions. \textcolor{black}{It is shown in \cite{zhang2021vinvl} that the usage of VinVL features and training on multiple datasets can significantly improve the performance of the original OSCAR on a variety of downstream Natural Language Processing (NLP) tasks, and therefore the learned Vision-language fusion model is named as OSCAR+.}
VinVL model \cite{zhang2021vinvl} has achieved the state-of-the-art performance in VL tasks, \textcolor{black}{and the performance of its proposed OSCAR+ has also surpassed that of others on downstream NLP tasks.} 

Inspired by recent progress in VL models and the requirements of semantic understanding in \textit{emotionship} analysis, we have adopted VinVL \cite{zhang2021vinvl} and \textcolor{black}{its proposed OSCAR+ \cite{zhang2021vinvl}} in \textit{EMOShip}. The benefits of using VinVL \textcolor{black}{and OSCAR+} are \textcolor{black}{threefold}. First, the semantic attributes of RoIs can be predicted, which can be essential to achieve \textit{emotionship}-awareness. Another advantage is that the semantic features of RoIs are also provided in VinVL, and these semantic features have encoded sufficient sentimental clues that can be fused together with eye-expression-related information to achieve more accurate emotion prediction. \textcolor{black}{Last but not least, we are able to perform language analysis tasks like Question Answering (QA) through using OSCAR+, which allows our eyewear system to capture the summary tag of a visual region.}
To the best of our knowledge, we are the first to integrate VL \textcolor{black}{and NLP} model into an eyewear system to enable the awareness of semantic attributes. 

\subsection{Gating Mechanisms}
Gating mechanism \footnote{Note that gating mechanism is more commonly known as \textit{attention mechanism}, however, to avoid confusion with the visual attention concept, we use the term \textit{gating mechanism} in this work} \cite{bahdanau2014neural, luong2015effective, vaswani2017attention, wang2018non} is an approach of spending more resources on those more informative parts of the input data. Typically, for an input signal, the importance degree for each of its position is weighted through a gating model and the output will be a signal with properly enlarged or shrunk values. There are a variety of gating models like the transformer \cite{vaswani2017attention} and the non-local network \cite{wang2018non}, and they are widely utilized in different fields like lip-reading \cite{ma2021lip} and image captioning \cite{xu2015show}. Among those gating models, Squeeze-and-Excitation (SE) network \cite{hu2018squeeze} is most closely related to our smart glasses \textit{EMOShip}. For a deep feature, SE network can learn the pattern of importance degree in a channel-wise manner, generating a scaling vector that adjusts each feature channel. In \textit{EMOShip}, SE is employed when fusing the semantic features from VL models and eye features to predict the emotional state, and more importantly, to learn the emotional impacts from scene images. 

\section{System Design}
\label{sctn::system}

This section presents \textit{EMOShip} system design. It first defines the \emph{emotionship} analysis problem, highlights the corresponding challenges, and then presents \emph{EMOShip}-Net, the proposed deep \emph{emotionship} analysis network, and details \emph{EMOShip} system software-hardware design and operation. 

\subsection{Problem Definition}
\label{sec::problem_definition}
Emotion recognition methods for eyewear devices aim to identify the emotional state from expressions (typically using the eye images). There can be various criteria regarding the emotional state, and we adopt the widely-accepted standard following the works of \cite{ekman1992argument, wu2020emo}. Specifically, the emotion is discretely classified into six basic categories \cite{ekman1992argument} -- 
happiness, surprise, anger, fear, disgust and sadness. In addition, we employ neutrality to represent the absence of emotions as in \cite{wu2020emo}. Let 
$e^t \in \{0,1,2,3,4,5,6\}$ represent the emotional state at time step $t$
and let $\mathbf{E}^t \in \mathbb{R}^{H_1 \times W_1 \times 3}$ be the eye images with height $H_1$ and width $W_1$. Recent smart eyewear devices \cite{wu2020emo, hickson2019eyemotion} utilized a deep network $\mathcal{N}_{eye}$ to obtain emotional predictions from eye images, i.e., $e^t=\mathcal{N}_{eye}(\mathbf{E}^t)$. 

In this work, we aim to solve the \textit{emotionship} analysis problem for eyewear devices, a task that is related to emotion recognition but is more sophisticated and ambitious.  
In particular, the emotional state is learned from both eye images and visual perceptions, and the impacts of visual perceptions on this emotional state, i.e., \textit{emotionship}, should also be quantitatively evaluated. Since the visual attentive region usually covers multiple semantic objects, the semantic attributes of the visual perceptions should be distinguished to avoid confusion of those objects. 

Let $\mathbf{I^t} \in \mathbb{R}^{H_2 \times W_2 \times 3}$ represent the scene image with height $H_2$ and width $W_2$, the user's visual attentive region is the priority to determine. In other words, we need to know which part of the scene image is attended by the user, which is formally known as Region of Interest (RoI).
We denote this RoI as $\mathbf{r}^t \in \mathbb{R}^{4}$, and $\mathbf{r}^t$ can be described as a rectangular area $(x_r^t, y_r^t, w_r^t, h_r^t)$, where $(x_r^t, y_r^t)$ is the central point of the rectangle, and $(w_r^t, h_r^t)$ denote the width and height of the rectangle, respectively, i.e., $\mathbf{r}^t \in \mathbb{R}^{4}$. The visual perceptions, denoted as $\mathbf{I}_{att}^t$, are obtained by cropping the region $\mathbf{r}^t$ out of $\mathbf{I}^t$. Different from regular emotion recognition methods, we aim to determine the emotional state $e^t$ from both the visual perceptions $\mathbf{I}_{att}^t$ and the eye image $\mathbf{E}^t$ through a deep model $\mathcal{N}_{1}$, i.e., $e^t=\mathcal{N}_{1}(\mathbf{I}_{att}^t, \mathbf{E}^t)$. 

Besides $e^t$, we also want to measure the impacts of visual perceptions $\mathbf{I}_{att}^t$ on this emotional state, that is, to which degree can this emotion be attributed to the visual experiences. We define this impact as an influential score $IS^t \in [0,1)$ which can be computed by 
inferring from $\mathbf{I}_{att}^t$, $\mathbf{E}^t$ and $e^t$. Assuming a deep model $\mathcal{N}_2$ is utilized, this can be written as $IS^t=\mathcal{N}_2(\mathbf{I}_{att}^t, \mathbf{E}^t, e^t)$. Intuitively, a larger $IS^t$ score indicates that $e^t$ is more associated with what the user observes, and vice versa for a smaller value.

The awareness of emotional state $e^t$ and the influential score $IS^t$ is not sufficient to fully reveal the \textit{emotionship}, as we still need to understand the semantic attributes of visual attentions to unambiguously describe the potential cause for $e^t$. The semantic attribute is defined as a summary tag of the attentive region $\mathbf{I}_{att}^t$, e.g., \say{white, warm beaches} if $\mathbf{I}_{att}^t$ depicts a white beach in summer. We denote this summary tag as $\mathbf{s}^t$ and it clarifies the semantic cause for $e^t$ at an abstract level, which is typically overlooked in previous works. 

Let $\mathbf{ES}^t$ represent the \textit{emotionship} and it can be formulated as
\begin{equation}
\label{eq::emotionship}
\mathbf{ES}^t=(e^t, \mathbf{I}_{att}^t, \mathbf{s}^t, IS^t).
\end{equation}
Different from traditional emotion recognition that isolates user's emotional state from surroundings and only predicts  $e^t$, our \textit{emotionship} $\mathbf{ES}^t$ additionally encodes the potential causes for $e^t$, i.e., 
visual perceptions $\mathbf{I}_{att}^t$ with semantic attributes $\mathbf{s}^t$, while the degrees of their emotional influences are also indicated by $IS^t$. With the awareness of \textit{emotionship}, eyewear devices can understand the semantic causes for emotions and also learn how visual experiences can affect emotions in a personalized manner, which is highly desirable and attractive in wearable systems. However, there are a number of challenges. 

\subsection{Challenges}
\label{motivation&challenges}
To achieve \textit{emotionship} analysis, the challenges are threefold.

The first is how to appropriately discover the semantic attributes of visual attention. 
With the embedding of the forward-facing world camera, smart eyewear devices can already estimate the gaze points \cite{chang2021memx} using eye-tracking techniques. Gaze points can be a valuable guidance to track human attentions. However, knowing merely the gaze point is insufficient, as there can be multiple semantic objects near this point that can potentially lead to the current emotional state. To avoid ambiguity, we need to clearly identify the semantic meanings around the gaze point. 
In other words, the semantic summary tag $\mathbf{s}^t$ of the visual perceptions $\mathbf{I}_{att}^t$ is necessary, yet $\mathbf{s}^t$ can be challenging to obtain, especially for eyewear devices.
In this work, we take inspirations from recent progress in visual features models \cite{zhang2021vinvl} to extract the tag $\mathbf{s}^t$, as detailed in Section \ref{workflow}. 

After the visual attentive regions have been located with semantic understandings, another challenge is how to establish the associations of human visual attention with the emotional state. The reason for emotion alternations can be subtle and difficult to determine. It could be highly associated with sentimental visual perceptions, e.g., when a user observes that a child is playing with parents and then cheers up, we can reasonably assume that the happiness is caused by this scene.
However, in cases that sentimentally-neutral visions are captured, 
a sudden change of emotional signals does not necessarily link with them. Therefore, it is crucial to correctly identify visual attentions' emotional contribution, i.e., to compute its influential score $IS^t$. 
In our workflow, $IS^t$ is automatically and implicitly learned with deep models, which is further described in Section \ref{workflow}. 

There is one more challenge when it comes to the prediction of emotional state $e^t$.  
Sentimental information in visual perceptions indeed provides insights on the potential cause of emotions, yet it is not reliable enough for recognizing emotion. Therefore, 
the utilization of expression-related information like eye-images in \cite{wu2020emo} is still indispensable. 
In other words, we infer the emotional state $e^t$ from both shreds of evidence, i.e., the sentimental clues in visual perceptions and the expression-related information in eye images,
leading to more robust emotion recognition performance. 
We have incorporated the Squeeze-and-Excitation network \cite{hu2018squeeze} to fuse them together, as described in Section \ref{workflow}.

To address the aforementioned challenges of \emph{emotionship} analysis, we have devised a deep network named \textit{EMOShip}-Net, the workflow of which is described in Section \ref{workflow}.

\begin{figure}[t!]
	\includegraphics[width=0.90 \textwidth]{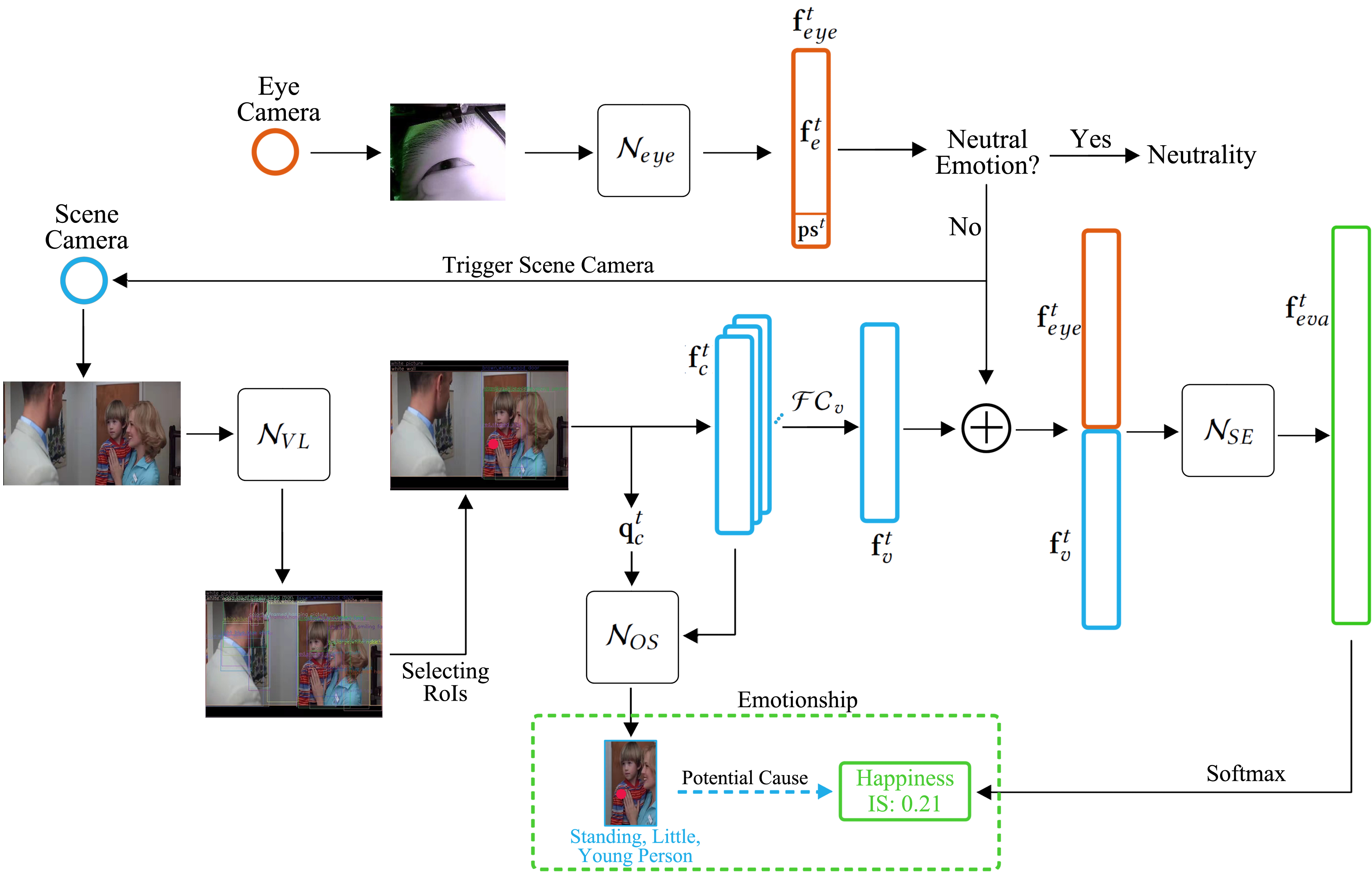}
 \caption{The workflow of \textit{EMOShip}-Net (Best Seen in Color).}
  	\label{fig::emoship_net}
\vspace{-1em}
 \end{figure}

\subsection{\textit{EMOShip}-Net Workflow}
\label{workflow}

\subsubsection{Overall Pipeline}
Fig. \ref{fig::emoship_net} illustrates the workflow of \textit{EMOShip}-Net.
At time step $t$, the input of the network contains two video streams: the eye images $\mathbf{E}^t$ taken by an inward-facing eye camera, and the scene images $\mathbf{I^t}$ recorded by another forward-facing world camera. The eye camera keeps tracking eye images $\mathbf{E}^t$ and monitors the rough emotional state, i.e., neutral or non-neutral. When a non-neutral emotion is spotted, the scene camera will be triggered to record scene $\mathbf{I^t}$. \textcolor{black}{A vision-language (VL) model ~\cite{zhang2021vinvl} is applied to extract all potential regions of interest (RoIs) with semantic tags in $\mathbf{I^t}$. The visual attentive region is determined from those RoIs based on the gaze point, and the summary tag for the selected area is obtained by a Question Answering (QA) process using the OSCAR+ \cite{li2020oscar} vision-language fusion model.} The features of the attentive regions, which are also provided by the VL model, are fused with the eye features using a Squeeze-and-Excitation (SE) network \cite{hu2018squeeze} to generate the final prediction on the emotional state. The scaling vector obtained after the SE network's excitation operation reveals a very important relationship, i.e., the emotional impact from visual attentions, or the influential score $IS^t$ as defined in Section \ref{sec::problem_definition}. 

\subsubsection{Extracting Eye Features}
Eye images $\mathbf{E}^t$ contain the information regarding expressions. \textcolor{black}{This work follows the EMO method \cite{wu2020emo} to extract expression-related features but makes necessary improvements to enhance the emotion recognition accuracy and suit our application scenarios. Specifically, EMO \cite{wu2020emo} consists of a feature extractor for eye images and a customized emotional classifier. Since emotion recognition in eye images is not the major pursuit of this work, we only adopt the former (feature extractor based on ResNet-18 backbone \cite{he2016deep} which is denoted as $\mathcal{N}_{eye}$) to extract $\mathbf{f}_{e}^t \in \mathbb{R}^{128}$, i.e. $\mathbf{f}_{e}^t=\mathcal{N}_{eye}(\mathbf{E}^t)$ but replace the latter one (the customized classifier) with a binary classifier for neutral/non-neutral predictions. More importantly, we have appended pupil information to $\mathbf{f}_{e}^t$ before feeding it into the binary classifier.}
This is inspired by \cite{aracena2015neural}, a work showing that statistical eye information such as pupil size can help to improve the emotion recognition accuracy. Denoting the pupil size information as $\mathbf{ps}^t \in \mathbb{R}^{2}$, we treat $\mathbf{ps}^t$ as an expert information, and following \cite{wang2020dynamic, Zhao2021A}, we concatenate this expert information $\mathbf{ps}^t$ with $\mathbf{f}_{e}^t$, which can be written as
$\mathbf{f}_{eye}^t=[\mathbf{f}_{e}^t, \mathbf{ps}^t]$, where the square bracket indicates the channel-wise concatenations.
Eye features $\mathbf{f}_{eye}^t  \in \mathbb{R}^{130}$ encode the expression-related information   
within eye regions and can be seen as an effective emotional indicator.
Note that $\mathcal{N}_{eye}$ will only be applied to eye images when a particular eye attention pattern  \cite{chang2021memx} is spotted to save energy, which will be detailed in Section \ref{sec::energy_software}. 
The trigger of the world camera, on the other hand, depends on eye feature $\mathbf{f}_{eye}^t$.

\subsubsection{Triggering the World Camera}
The high-resolution world camera is more energy-intensive than the low-resolution light-weight eye camera, thus it would be too energy costly to capture scene frames nonstop. Considering the energy limitations of wearable devices, we have designed a \say{trigger} mechanism for the world camera. The idea is to skip those emotional-neutral frames (there is no need to analyze the \textit{emotionship} for neutral emotions) 
and focus on those frames with non-neural emotions. In particular, we design a binary classifier $\mathcal{C}_{eye}$ to separate $\mathbf{f}_{eye}^t$ into neutral or non-neutral emotion, respectively. 
If $\mathbf{f}_{eye}^t$ is believed to fall into the neutral-emotional category,
the world camera will be disabled to avoid unnecessary energy costs. 
Otherwise, it will be triggered to enable the following operations. 

\subsubsection{Selecting RoI candidates}
\label{sssec:select_roi_candidates}
The triggered forward-facing world camera records the scene images $\mathbf{I}^t$. 
Using the eye-tracking technique in \cite{kassner2014pupil}, we first estimate from $\mathbf{E}^t$ the gaze point $\mathbf{g}^t=(x_g^t, y_g^t)$ where $x_g^t$ and $y_g^t$ refer to the 2D coordinates of this gaze point with respect to scene image $\mathbf{I}^t$. Since $\mathbf{g}^t$ is a 2D point, we still need to find the region of visual perceptions $\mathbf{r}^t=(x_r^t, y_r^t, w_r^t, h_r^t)$. In this work, we use the VinVL model \cite{zhang2021vinvl} to generate all potential regions in $\mathbf{I}^t$, and then we perform a filtering process to select certain RoI candidates for $\mathbf{r}^t$ from all those regions.

In particular, denote the VinVL model \cite{zhang2021vinvl} as $\mathcal{N}_{VL}$. Given the scene image $\mathbf{I}^t$, $\mathcal{N}_{VL}$ is able to generate a total of $K$ potential regions $\{\mathbf{R}_1^t, \mathbf{R}_2^t, \ldots, \mathbf{R}_K^t\}$ where $\mathbf{R}_i^t \in \mathbb{R}^{4}$ represents the $i$-th candidate. 
Note that for a RoI $\mathbf{R}_i^t$, its corresponding visual
representation (or feature) $\mathbf{f}_{R_i}^t \in \mathbb{R}^{2048}$ and the semantic representation $\mathbf{q}_{R_i}^t$ (e.g., a tag) is already given by $\mathcal{N}_{VL}$. 
We have designed a filter process to select the ten most suitable regions out of all those $K$ regions based on the gaze point $\mathbf{g}^t$. 
That is, for each candidate $\mathbf{R}_i^t$, we compute the Euclidean distance from its central point to the gaze point $\mathbf{g}^t$. 
Then we empirically select the top ten regions with the smallest distances, i.e., the ten regions that are closest to the gaze point. 
Denoted those regions as $\mathbf{R}_{c}^t=\{\mathbf{R}_{c1}^t, \mathbf{R}_{c2}^t, ..., \mathbf{R}_{c10}^t\}$, they are the most relevant RoIs with the visual attentive region. 
After this filtering process, there are still ten RoI candidates, and we need to determine a final visual attention region and also to generate its summary tag. To achieve this, we perform two Question Answering (QA) sessions. 

\subsubsection{Determining Visual Attentive Region and Summary Tag}
\label{sssec:var}
 Recall that we have already selected ten candidates of regions $\mathbf{R}_{c}^t=\{\mathbf{R}_{c1}^t, \mathbf{R}_{c2}^t, \ldots, \mathbf{R}_{c10}^t\}$, and for the $i$-th region $\mathbf{R}_{ci}$, its visual feature $\mathbf{f}_{ci}^t \in \mathbb{R}^{2048}$ and its semantic representation $\mathbf{q}_{ci}^t$ are already provided by $\mathcal{N}_{VL}$. 
 To select the actual visual attentive region and to generate it summary tags, we perform a Visual Question Answering (VQA) \cite{goyal2017making} and an Image Captioning \cite{aneja2018convolutional}, \textcolor{black}{based on the OSCAR+ vision-language fusion model \cite{zhang2021vinvl}.} Specially, 
 denote the visual features of selected ten regions as $\mathbf{f}_{c}^t=\{\mathbf{f}_{c1}^t, \mathbf{f}_{c2}^t, \ldots, \mathbf{f}_{c10}^t\}$, denote the semantic attributes as $\mathbf{q}_{c}^t=\{\mathbf{q}_{c1}^t, \mathbf{q}_{c2}^t, \ldots, \mathbf{q}_{c10}^t\}$, \textcolor{black}{and denote the OSCAR+ model \cite{zhang2021vinvl} as $\mathcal{N}_{OS}$}. 
 The VQA session aims to determine the appropriate visual attentive region $\mathbf{r}^t$. First we invoke $\mathcal{N}_{OS}$ to answer the question ${\mathbf{Q}_1}$ \say{What object makes people feel happy/surprised/sad/angry/feared/disgusted?} by also inferring to $\mathbf{f}_{c}^t$ and $\mathbf{q}_{c}^t$ and \textcolor{black}{obtain an answer $\mathbf{a}^t$ from $\mathcal{N}_{OS}$},
 which written as
 \begin{equation}
\label{eq::final_roi}
\mathbf{a}^t=\mathcal{N}_{OS}(\mathbf{Q}_1, \mathbf{f}_{c}^t, \mathbf{q}_{c}^t).
\end{equation}
\textcolor{black}{Then among the ten attributes $\mathbf{q}_{c}^t$, we find the one $\mathbf{q}_{cj}^t$ whose word2vec embedding \cite{pennington2014glove} is closest to that of answer $\mathbf{a}^t$ than all other attributes, and $\mathbf{q}_{cj}^t$'s corresponding region $\mathbf{R}_{cj}^t$ is seen as the visual attentive region $\mathbf{r}^t$, i.e. $\mathbf{r}^t = \mathbf{R}_{cj}^t$.} 

As for the Image Captioning (IC) session, the target is to generate an appropriate tag that summarize the semantic attributes of visual perception region. In this session, there is no question to answer, \textcolor{black}{and $\mathcal{N}_{OS}$ only looks at the visual features $\mathbf{f}_{c}^t$ to generate the tags, which can be expressed as}
 \begin{equation}
\label{eq::summary_tag}
\mathbf{s}^t=\mathcal{N}_{OS}(\mathbf{f}_{c}^t)
\end{equation}
where $\mathbf{s}^t$ is the summary tag for visual attentive region. 

\subsubsection{Determining the Emotional State}
The emotional state $e^t$ is obtained through a synthesis analyze of the eye feature $\mathbf{f}_{eye}^t$ and the visual features $\mathbf{f}_{c}^t=\{\mathbf{f}_{c1}^t, \mathbf{f}_{c2}^t, \ldots, \mathbf{f}_{c10}^t\}$ of the candidate regions. 
Particularly, since $\mathbf{f}_{c}^t \in \mathbb{R}^{10 \times 2048}$ and $\mathbf{f}_{eye}^t \in \mathbb{R}^{130}$, we first employ a Fully Connected (FC) layer  $\mathbf{f}_{c}^t$ to summarize the visual attributes and reduce its dimensionality, i.e., $\mathbf{f}_{v}^t=\mathcal{FC}_{v}(\mathbf{f}_{c}^t)$
where $\mathcal{FC}_{v}$ denotes the FC layer and $\mathbf{f}_{v}^t \in \mathbb{R}^{130}$. 
We concatenate the channels of visual perceptions' feature $\mathbf{f}_{v}^t$ and eye feature $\mathbf{f}_{eye}^t$ to formulate $\mathbf{f}_{ev}^t=[\mathbf{f}_{v}^t, \mathbf{f}_{eye}^t]$. This concatenated feature $\mathbf{f}_{ev}^t \in \mathbb{R}^{260}$ contains emotional evidences from both the eye and scene images, and it is 
fed into a Squeeze-and-Excitation (SE) network \cite{hu2018squeeze} $\mathcal{N}_{SE}$ to obtain a scaling vector $\mathbf{u}^t \in \mathbb{R}^{260}$, i.e. $\mathbf{u}^t=\mathcal{N}_{SE}(\mathbf{f}_{ev}^t)$. This scaling vector $\mathbf{u}^t$ is channel-wisely multiplied with the the concatenated features $\mathbf{f}_{ev}^t$ to obtain feature $\mathbf{f}_{eva}^t \in \mathbb{R}^{260}$, i.e. $\mathbf{f}_{eva}^t=\mathbf{u}^t * \mathbf{f}_{ev}^t$ where $*$ represents the channel-wise multiplication. 
Note that the scaling vector $\mathbf{u}^t$ is learned from the SE gating mechanisms and it reflects the importance degree of each channel in $\mathbf{f}_{ev}^t$. 
The obtained feature $\mathbf{f}_{eva}^t$ is then input into a soft-max classifier $\mathcal{C}_{EMO}$ to generate the final emotion prediction $e^t$, i.e. $e^t=\mathcal{C}_{EMO}(\mathbf{f}_{eva}^t)$. 

\subsubsection{Computing the Influential Score}
The Influential Score indicates the degree of emotional impacts from visual perceptions, and it can be computed from the scaling vector $\mathbf{u}^t$ learned from SE gating mechanism. Recall that $\mathbf{u}^t$ represents the importance degree of each channels in $\mathbf{f}_{ev}^t$, while $\mathbf{f}_{ev}^t$ is concatenated from eye features $\mathbf{f}_{eye}^t$ and visual perception's feature $\mathbf{f}_{v}^t$. We are therefore able to evaluate the importance degree of visual perception's feature $\mathbf{f}_{v}^t$ in predicting emotional state, or the Influential Score $\mathit{IS}$, through the usage of $\mathbf{u}^t$, which can be written as 
\begin{equation}
\label{eq::IS}
{\mathit{IS}}^t=\frac{\sum_{i=1}^{130}{\mathbf{u}_i^{t}}} {\sum_{i}{\mathbf{u}_i^{t}}}
\end{equation}
where $\mathbf{u}_i^{t}$ denotes the $i$-th scalar of $\mathbf{u}^{t}$, and we assume the first 130 scalars of $\mathbf{u}^{t}$ corresponds to channels of $\mathbf{f}_{v}^t$. 
Using the influential scores in Equation \ref{eq::IS}, we can determine to which degree an emotional state was affected by the sentimental visions. For instance, if a really small ${IS}^t$ is computed, we could conclude that the current emotional status is not related to the observed visual perceptions. In contrast, a large ${IS}^t$ value implies that the current emotion is highly related to the attentive scene regions.  

\subsubsection{Emotional State in Video Sequence}
To maintain simplicity, we only illustrate how to predict emotion for a certain time step $t$ in the descriptions so far.  However, emotions are temporally consistent processes instead of static ones, and therefore we also need to consider how to aggregate the emotion prediction of different time steps. For an emotion video clip of $T$ frames, assuming that we have already computed emotion prediction for each frame, i.e. $\{e^1,e^2,...e^T\}$, we use the majority emotion class $e_{m}$ as the emotion prediction of this sequence. Formally, the majority emotion class $e_{m}$ can be computed as 
\begin{align*} 
e_m=\argmax_{i}\sum_{j=1}^{T}{\mathbbm{1}(e^j=i)} 
\end{align*} 
where $i \in \{0,1,2,3,4,5,6\}$, $\mathbbm{1}$ represents the indicator function, $\mathbbm{1}(e^j=i)=1$ if $e^j=i$ and $\mathbbm{1}(e^j=i)=0$ if $e^j\neq i$. 

\subsection{\textit{EMOShip} System}

\subsubsection{Hardware Design}
The \textit{EMOShip} prototype is a smart eyewear system equipped with two cameras, including one outward-facing world
camera and one customized inward-facing eye camera. The outward-facing world camera collects the visual content 
aligned with the wearer's field of view. We adopt Logitech B525 1080p HD Webcam (1280$\times$960@30fps).
The inward-facing eye camera supports continuous eye tracking and eye-related expression 
feature capture. We adopt a camera module with GalaxyCore GC0308 sensor (320$\times$240@30fps), equipped with an IR LED light to illuminate the iris. 
\textcolor{black}{This hardware module is inspired by the Pupil \footnote{\url{https://pupil-labs.com}}, but we have re-implemented it accordingly to suit our needs. }
The \textit{EMOShip} prototype is equipped with Qualcomm's wearable system-on-module 
solution, combining Qualcomm Snapdragon XR1 chipset with an eMCP(4GB LPDDR/64GB eMMC). Its typical power consumption 
is \unit[1]{W}, which is suitable for battery-powered wearable design. 

\subsubsection{Software Operation}
\label{sec::energy_software}
In \textit{EMOShip}, \textit{EMOShip}-Net performs emotion recognition and \textit{emotionship} analysis. Targeting energy-constrained wearable scenarios, \textit{EMOShip}-Net is equipped with a carefully designed energy-efficient workflow as follows. 

\begin{itemize}
\item First, \textit{EMOShip}-Net continuously senses the eye camera to perform eye tracking. 
To minimize the energy cost of eye tracking, \textit{EMOShip}-Net uses a computationally efficient pupil detection 
and tracking method~\cite{kassner2014pupil} to detect potential attention events. Following the work by Chang et al.~\cite{chang2021memx}, 
a potential attention event needs to satisfy two conditions simultaneously: (1) there is a temporal transition from saccade to 
smooth pursuit, which suggests a potential visual attention shift; and (2) the gaze follows a moving target or fixating on a 
stationary target. \textcolor{black}{We modify the open-source Pupil software \footnote{\url{https://github.com/pupil-labs/pupil/releases}} to achieve more accurate eye movement pattern detection that can better satisfy the requirement of our system. Pupil software predicts two eye movements---fixation and non-fixation, based on the degree of visual angles \cite{kassner2014pupil}. However, when we deployed it in our system, we needed more eye movements such as saccade and smooth pursuit to detect a potential visual attention event. To address this issue, we follow the method proposed in~\cite{chang2021memx} and leverage the historical gaze trajectory profile to give a more accurate eye movement detection. The behind intuition is that possible smooth pursuit or fixation eye movements occur when the historical gaze points are located in a spatially close region for a while; otherwise, a saccade eye movement occurs. What we want is a stable detection of eye movements, and therefore we pay more attention to the recall of our method.
The overall recall is \textcolor{black}{99.3\%}, which shows our eye movement detection method is working robustly and reliably. Also, the} inference time of the eye-tracking method used in \textit{EMOShip}-Net is \unit[115.1]{fps}, 
or \unit[8.7]{ms/frame}.

\item Once a potential attention event is detected by the computationally efficient eye-tracking method, \textit{EMOShip}-Net takes the eye images as the input of the light-weight network $\mathcal{N}_{\mathit{eye}}$ to extract eye-related expression-related information 
and performs neutral vs. non-neutral emotional state classification. $\mathcal{N}_{\mathit{eye}}$ is computationally efficient, which
only requires \unit[20.3]{ms} to perform emotional state classification for each eye image frame.

\item Only when a non-neutral emotional state is detected, \textit{EMOShip}-Net turns on the high-resolution world camera 
to sense the scene content for semantic analysis. In other words, the high-resolution (more energy expensive) scene content 
capture and processing 
the pipeline remains off most of the time, avoiding unnecessary data sensing and processing, thereby effectively improving the overall 
system energy efficiency.  

\item Finally, \textit{EMOShip}-Net leverages a cloud infrastructure to perform computation-intensive semantic attribute feature
extraction and eye-scene feature aggregation to support final emotion recognition and \textit{emotionship} analysis, thus avoiding the  
energy consumption on the eyewear device side. 

\end{itemize}

The energy consumption of the \textit{EMOShip} eyewear device is estimated  as follows:
\begin{align}
	E_{\mathit{\textit{EMOShip}}}&= T_{\mathit{always-on}} \times (P_{\mathit{eye \;camera}} + P_{\mathit{eye\; tracking}})  + T_{\mathcal{N}_{\mathit{eye}}} \times P_{\mathcal{N}_{\mathit{eye}}}  + T_{\mathit{captured}} \times P_{\mathit{world\;camera}} ,
	\label{eq::energy_mdl}
\end{align} 
where $T_{\mathit{always-on}}$ is the overall operation time of the \textit{EMOShip} eyewear device, $P_{\mathit{eye\; camera}}$ and $P_{\mathit{world\; camera}}$ is the power consumption of the eye camera and the world camera, respectively, \textcolor{black}{$T_{\mathcal{N}_{\mathit{eye}}} $ is the operation time of the light-weight eye analysis network $\mathcal{N}_{\mathit{eye}}$, }
and $T_{\mathit{captured}} $ is the operation time of the high-resolution video recording when non-neutral emotional states are detected.

\begin{figure}[!h]
	\flushright
	\includegraphics[width=1.0 \textwidth]{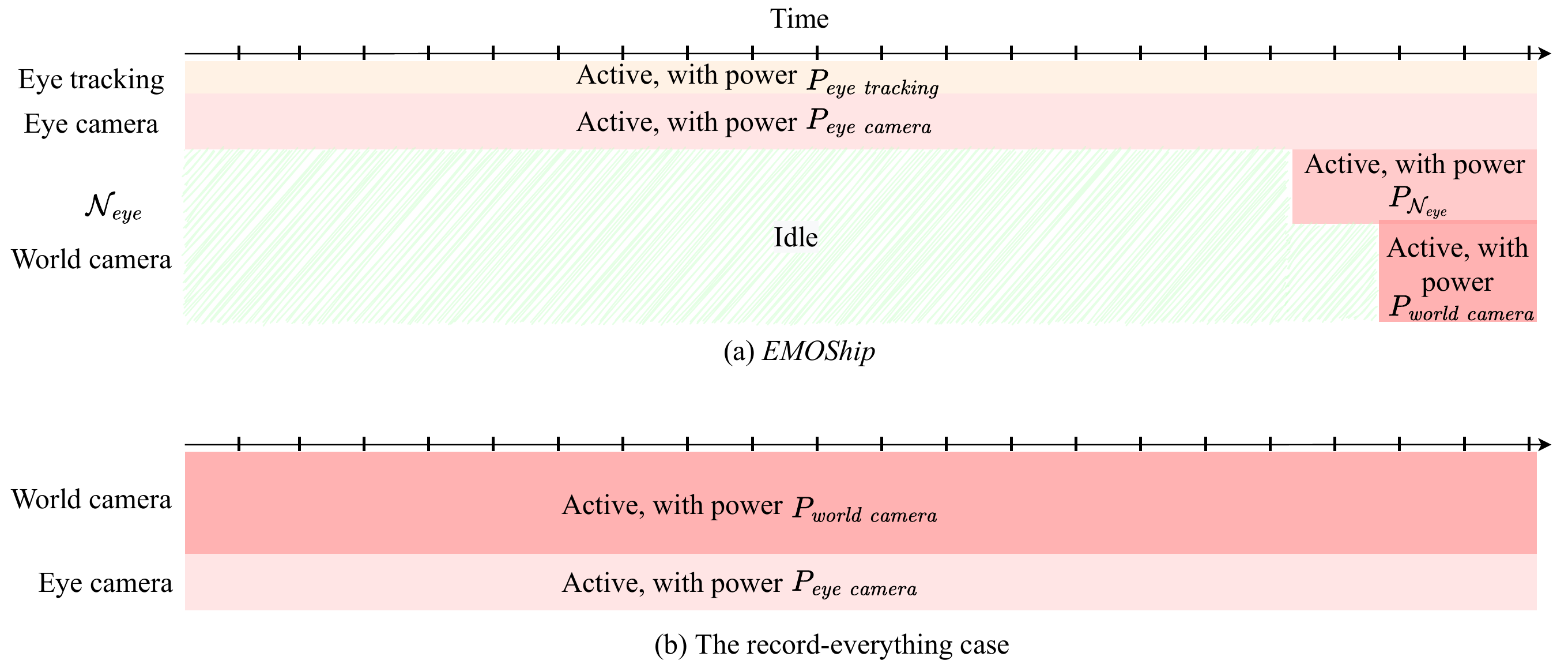}
	\caption{The run-time operation of \textit{EMOShip} (a), as well as that of the recording-everything case (b).}
	\label{fig::power}
\end{figure}

\Note{*Yuhu: double check this paragraph*}
Fig.~\ref{fig::power}(a) illustrates the run-time operation of \textit{EMOShip} on the eyewear. 
Targeting the Qualcomm Snapdragon wearable platform, physical measurement shows that $P_{\mathit{eye \; camera}} = 0.07W$, $P_{\mathit{eye\;tracking}} = 0.1 W$, $P_{\mathit{world \; camera}} = 1.3 W$, and $P_{\mathcal{N}_{\mathit{eye}}} = 1.1 W$. 
Targeting the real-world pilot studies described in Section~\ref{sctn::app}, physical measurement shows that $\mathcal{N}_{\mathit{eye}}$ and the world camera remain off during \textcolor{black}{86.8\% and 94.6\%} of the system operation time, respectively.
We can estimate that, considering a \unit[2.1]{Wh} battery (similarly to Google Glass Explorer Edition), \textit{EMOShip} can support \unit[5.5]{hours} continuous operation, which can meet typical daily usage needs without frequent charging.

For comparison purposes, let's consider the record-everything case (shown in Figure~\ref{fig::power}(b)), with both eye camera and world camera remaining on all the time. In this case, the overall system energy consumption is $E_{\mathit{always-on}} = T_{\mathit{always-on}} \times (P_{\mathit{eye \;camera}} 
+ P_{\mathit{world\; camera}})$, and the system battery lifetime is  approximately \unit[1.5]{hours}.
Compared with the record-everything case, \textit{EMOShip} improves the system battery lifetime by 3.6X. 
\section{Evaluation}
\label{sctn::exp}
This section presents the in-lab experiments to evaluate the performance of \textit{EMOShip}. 

\subsection{Dataset}  
To evaluate \textit{EMOShip}, we need the scene images observed by the wearer, the wearer's eye images, and also the emotional states during this observation process. In other words, an eligible dataset should cover both the scene and eye timelines and also contain emotion annotations of the same duration. However, most publicly-available emotion datasets cannot satisfy those requirements, since they either lack the scene images like the MUG dataset~\cite{aifanti2010mug}
or do not provide the facial or eye regions such as the FilmStim dataset~\cite{schaefer2010assessing}. 
Therefore, we collect and build a new dataset named EMO-Film to suit our needs, which is \textcolor{black}{available online at \footnote{\url{https://github.com/MemX-Research/EMOShip}}} and detailed below. 

\subsubsection{Data Collection}
\label{sstn::data_collection}
\begin{figure}[h]
	\includegraphics[width=0.98 \textwidth]{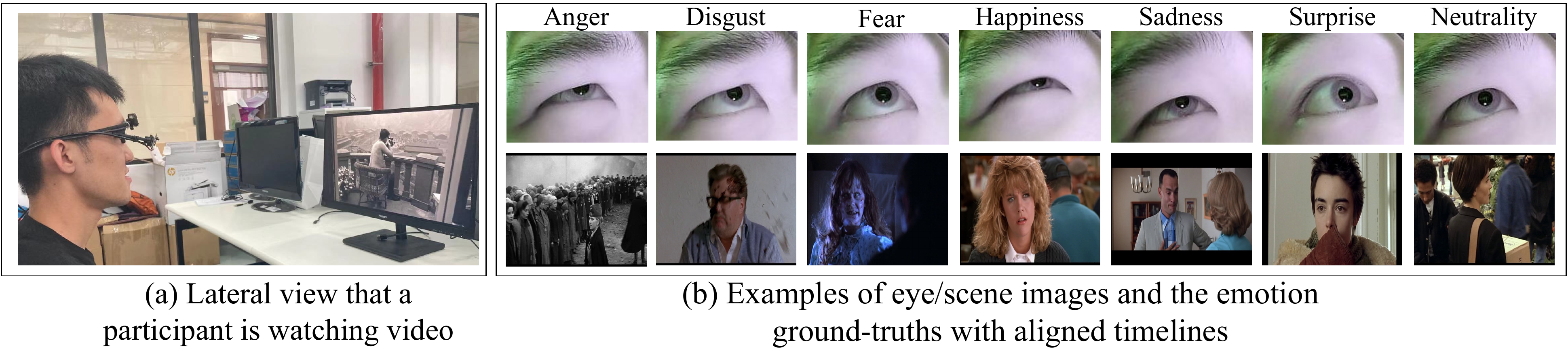}
 \caption{Data collection of EMO-Film: The laboratory setting (left), and the eye and scene images of seven emotional states from the same participant (right).}
 	\label{fig::emotion}
 \end{figure}

The data of EMO-Film dataset is collected in a controlled laboratory environment. As shown in Fig.~\ref{fig::emotion} (left), participants \textcolor{black}{equipped with \textit{EMOShip} was instructed to} watch several emotion-eliciting video clips displayed on a desktop monitor. A total of 20 volunteers attended the data collection of EMO-Film, including 8 females and 12 males.

\paragraph{(1) Video Data Preparation.}
The video clips were selected from the FilmStim dataset  \footnote{\url{https://nemo.psp.ucl.ac.be/FilmStim/}} \cite{schaefer2010assessing} \textcolor{black}{as FilmStim is one of the widely-used emotion-eliciting video dataset~\cite{wu2020emo}}. \textcolor{black}{We first divide all videos of FilmStim dataset  (\textcolor{black}{64} video clips in total) into \textcolor{black}{7} categories based on the provided sentiment labels, each category corresponding to one emotional class (the neutral plus six basic emotion). 
Then we randomly sample at least one video clip from each category summing up to \textcolor{black}{6-7} for a participant to watch, which may take approximately \textcolor{black}{\unit[20]{minutes}} to complete.  
Note that the film clips in FilmStim dataset are already sentimentally sufficient to evoke emotional reactions at certain degrees, and such design can ensure that all the six basic emotions will be evenly covered. We also ensure that each film clip was watched by at least \textcolor{black}{two} subjects.  }

\paragraph{(2) Data Collection Process.}
During the watching process, we kept recording the eye regions of participants using the eye camera. To ensure the video scenes can be captured properly, we pre-adjusted the field of view of the world camera to be aligned with the monitor and recorded the displayed video simultaneously. 
In this way, we are able to gather the eye/scene data and the emotion ground-truths with aligned timelines, as shown in Fig.~\ref{fig::emotion} (right).  \textcolor{black}{This recording session can typically take around  \textcolor{black}{\unit[20]{minutes}} per people.}

\paragraph{(3) Labeling Process.}
\textcolor{black}{After all the scheduled movie clips displayed, the participant takes a short break (around \textcolor{black}{20} minutes) and then is instructed to label their emotional states. This labeling process can take up to \textcolor{black}{\unit[70]{minutes}} (compared with \textcolor{black}{20} minutes of watching the films) and the generated emotional annotations are arguably accurate since the videos were just shown around \textcolor{black}{20} minutes ago.
We develop a labeling tool with a GUI window to facilitate this process. Each participant is orally told how to use this tool, i.e. for each eye/scene image pair, the participant indicates their emotional state by clicking on the corresponding button or using a keyboard shortcut. There are a total of seven emotional states to choose from: neutral plus the six basic emotions. We consider only one emotional state per time instant for simplicity. This process is repeated until all eye/scene image pairs have been assigned labels.}

The whole data collection process takes approximately four days, and the gathered data and labeling last for approximately 1.5 hours per participant.

\subsubsection{Dataset Statistics} 
EMO-Film dataset is further divided into two sets for the purpose of training/testing, respectively. 
\textcolor{black}{We split the video data of each subject into 80\%/20\% for training/testing based on the timestamps. 
The 80\% clips with smaller timestamps (i.e. recorded at an earlier time) are assigned as the train set, and the rest 20\% clips as the test set. 
The overall percentages of video sequences belong to \say{anger}/\say{disgust}/\say{dear}/\say{happiness}/\say{sadness}/\say{surprise} are 2.9\%/18.2\%/20.8\%/20.0\%/20.8\%/17.3\%, respectively.
}

As shown in Table \ref{tb::trainset}, there are a total of 144,145/45,338 eye-scene image pairs
in the training/testing set, respectively. Each eye-scene frame pair is
properly aligned in timelines, and the frame-level emotion ground-truths is also provided. 
The resolution for scene image is 1280$\times$960, while that of eye images is 320$\times$240. 
The distribution of the seven emotion classes is also shown in Table \ref{tb::trainset}. As we can see, \say{fear} accounts for the largest non-neutral emotion, while the clips representing \say{anger} are the fewest. The number of \say{neutrality} clips is similar to that of \say{fear}.
\textcolor{black}{We also apply the data augmentation techniques, including the rotations, flips, and affine transforms, to balance the distribution of different emotional states during the training stage, which can be important to the training of \textit{EMOShip}-Net.}

\begin{table}[!htb]
\caption{Training/testing set distribution.} 
\label{tb::trainset}
\begin{tabular}{|c|c|c|c|c|c|c|c|c|}
\hline
Emotional States          & Anger & Disgust & Fear & Happiness & Sadness & Surprise & Neutrality \\ \hline
\begin{tabular}[c]{@{}c@{}}Number of eye-scene image pairs\\ in training set\end{tabular}    & 3,519   &  21,844        &   25,000    &  23,807     &  24,080   &  20,895    & 25,000          \\ \hline
\begin{tabular}[c]{@{}c@{}}Number of eye-scene image pairs\\ in testing set\end{tabular}     & 990  &  2,843       &   8,693    &  4,214     &  7,068   &  3,801  &  17,729     \\ \hline
\end{tabular}
\end{table}

\begin{figure}[!htb]
	\centering
		\includegraphics[width=.5 \textwidth]{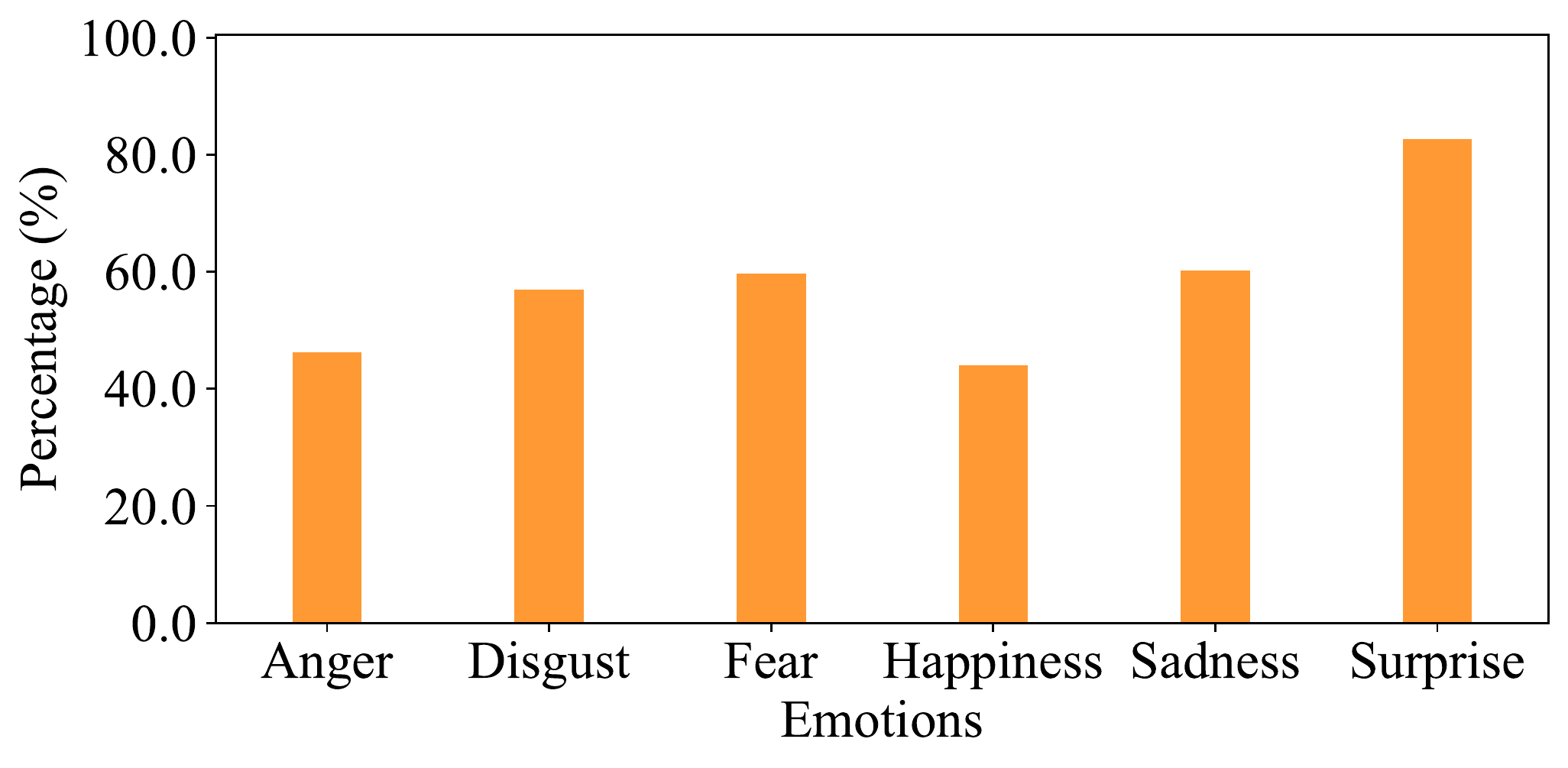}
	\caption{\textcolor{black}{The percentage of video frames where all subjects demonstrate exactly the same emotional reactions for different emotional classes.}}
		\label{fig::ratio}
\end{figure}

\textcolor{black}{When viewing identical sentimental contents, different participants may demonstrate different emotional reactions. This inter-participant variability is an interesting phenomenon to be examined and to be discussed. To examine this inter-participant variability, we first divide the videos into six sentimental categories excluding the neutrality, each category corresponding to one emotional class that they are likely to arouse, and then for each category, we calculate the percentage of video frames where all subjects demonstrate exactly the same emotional reactions. As shown in In Fig.~\ref{fig::ratio}, we can see that \say{surprise} can be most easily aroused and shared among people watching the same videos, while there are also a comparatively high proportion of subjects who share the same emotional feelings from viewing content related with \say{disgust}, \say{fear} and \say{sadness}. 
\say{Happiness} and \say{Anger}, however, have the lowest repeating probabilities, potentially because people have more personalized tastes on videos of those kinds. }

\subsection{Experimental Setup}
\subsubsection{Evaluation Metrics} 
The \textit{emotionship}, as defined in Equation \ref{eq::emotionship}, depicts the emotional state of the users, and also describes the potential cause of such emotions. Since the frame-level ground-truths of emotions are already provided in our EMO-Film dataset, the evaluation of the former (emotional state prediction) is comparatively straight-forward. Following \cite{2014A}, we adopt the multilabel-based macro-averaging metric to evaluate the performance of emotional state predictions, as define in Equation~(\ref{eq::macro_avg}).
\begin{equation}
\label{eq::macro_avg}
B_{\mathit{macro}}(h) = \frac{1}{C}\sum_{j=1}^{C} B(\mathit{TP}_{j},\mathit{FP}_{j},\mathit{TN}_{j},\mathit{FN}_{j}),
\end{equation}
where $B(\mathit{TP}_{j},\mathit{FP}_{j},\mathit{TN}_{j},\mathit{FN}_{j})$ represents binary classification performance on label $j$ ($B \in \{\mathit{Accuracy},\mathit{Precision}, \mathit{Recall}\}$). $C$ is the number of emotion classes, in this study, $C=6$. That is, we only recognize the six non-neutral emotions. $\mathit{TP}_{j}$, $\mathit{FP}_{j}$, $\mathit{TN}_{j}$, and $\mathit{FN}_{j}$ denote the number of $\mathit{true\;positive}$, $\mathit{false\;positive}$, $\mathit{true\;negative}$, and $\mathit{false\;negative}$ test samples with respect to the $j$ class label, respectively.

However, as \textit{emotionship} itself is a new concept, there is no existing metric that can be used to evaluate the quality of the latter, i.e. the understanding of potential causes of emotions. Besides, it is also difficult to objectively annotate such a potential cause, as it is highly personalized, subjective and subtle. In this work, we follow an intuitive way of examining the quality of such understandings. Several representative samples are visualized to compare the qualities of semantic attributes generated by \textit{EMOShip} and a baseline based on VinVL model \cite{zhang2021vinvl}, and we also plot the varying trends of Influential Score from different scenarios to demonstrate that \textit{EMOShip} has correctly captured the emotional impacts from scene images.

\subsubsection{Baselines} 
\label{sec:baseline}
To evaluate the performance of emotion recognition, we have selected four works as baselines, which are: 1). the emotion-aware smart glasses EMO \cite{wu2020emo}, 2). EMO+ which is an improved version of EMO, 3). VinVL model \cite{wu2020emo} that extracts semantic scene features for emotion recognition, and 4). VinVL+ that is modified to focus on the attentive regions of users. 

\begin{enumerate}

 \item EMO \cite{wu2020emo} utilizes a deep CNN to recognize emotions from eye images and is closely related with our work, and therefore it is used as a primary baseline. Note that we have discarded the classifier in EMO as it requires the construction of an auxiliary face recognition database, which is resource-costly but only introduces very limited improvement. 

 \item Inspired by \cite{aracena2015neural}, we integrate the information of pupil sizes with EMO to improve its recognition accuracy. In particular, the pupil size of eyes is seen as a kind of expert information, and this expert information is concatenated to the second last Fully Connected (FC) layer of the CNN in a similar way with \cite{wang2020dynamic}. This baseline method is denoted as EMO+.

 \item Both EMO and EMO+ predict emotions from the eye images. However, hints on emotional states can also be fetched from scene images, especially from those sentimental visions that are more likely to evoke emotions. To validate this, we devise the third baseline method that only looks at the scene image and tries to predict emotional states from the sentimental clues in it. \textcolor{black}{In details, we utilize the VinVL model \cite{wu2020emo} to extract visual features from scene images containing sentimental information. Then, those visual features are fed to a classifier consisting of two layers to obtain the emotion predictions. Regarding the summary tag generation, all the visual features are input into the OSCAR+ model~\cite{zhang2021vinvl} to obtain summary tags.}
This approach is called VinVL for simplicity.

 \item The visual features of VinVL contain information from all potential Regions of Interest (RoIs). There can be various sentiment clues in those RoIs, however, it is the sentimental information from the user's attentive region that really matters. Therefore, we have set VinVL to focus on the user's visual attention and discard information from other irrelevant areas. This is achieved by only using visual features from the user's attentive region. This method is named VinVL+.
   
\end{enumerate}

As for the understanding of the potential cause of emotions, we compare summary tags generated by our \textit{EMOShip} with VinVL+ to provide an intuitive illustration of the qualities.

\subsubsection{Training \textit{EMOShip}-Net} 
The structure \textit{EMOShip}-Net is complicated, as it involves several backbone networks with significantly different architectures and design purposes. Instead of end-to-end training, we adopt an iterative way to learn \textit{EMOShip}-Net, i.e. each component network is trained individually while freezing the weights of other parts. 

\begin{figure}[htb]
	\includegraphics[width=0.8 \textwidth]{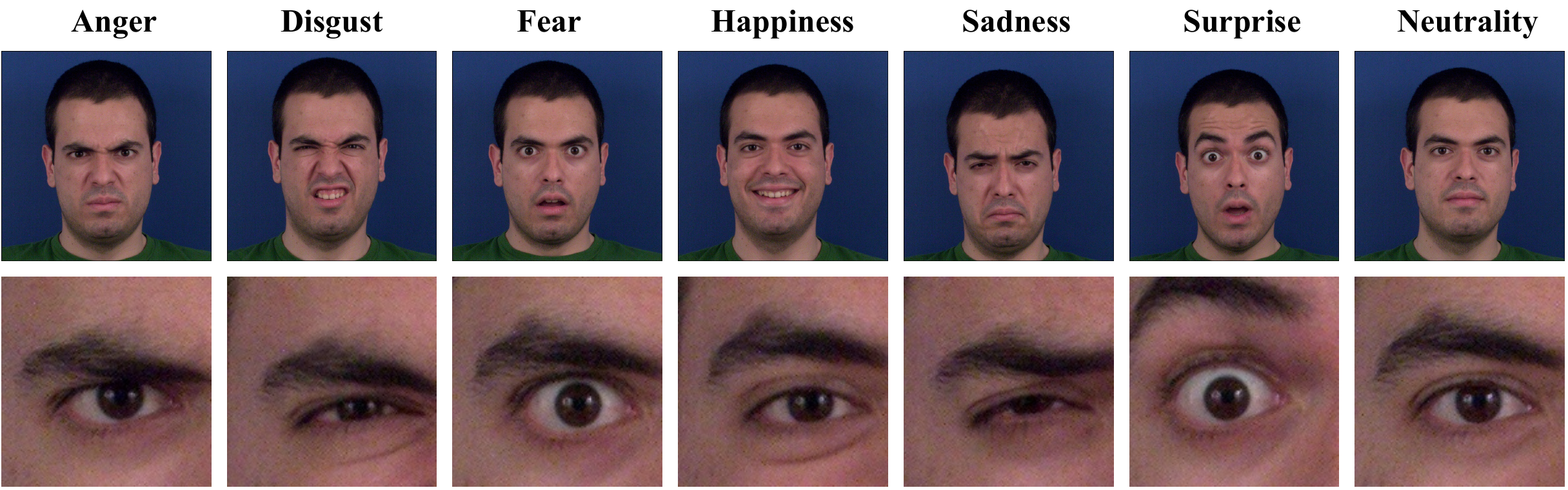}
 \caption{Seven emotional expressions of the original MUG facial expression examples (top row), our fine-tuning single-eye-area data cropped from MUG (bottom row). }
 	\label{fig::mug}
\end{figure}

The eye network $\mathcal{N}_{eye}$ is used frame-wisely and serves as the trigger for the scene camera, and therefore it is the first component to learn. We generally follow the training procedures in \cite{wu2020emo} and pre-train $\mathcal{N}_{eye}$ with cross-entropy loss on FER2013 dataset \cite{goodfellow2013challenges} and MUG dataset \cite{aifanti2010mug}. Note that on MUG dataset the eye regions are cropped out of the whole facial image, as shown in Fig. \ref{fig::mug}.
The pre-trained $\mathcal{N}_{eye}$ is further fine-tuned the training set of our collected EMO-Film dataset. 

Considering the high complexity in \textcolor{black}{visual features model} $\mathcal{N}_{VL}$ and the \textcolor{black}{vision-Language model} $\mathcal{N}_{OS}$, we directly utilize the pre-trained weights provided by authors of those two models. The Squeeze-and-Excitation model  $\mathcal{N}_{SE}$, is trained together with the FC layer $\mathcal{FC}_{v}$ on EMO-Film dataset. We use Adam \cite{kingma2014adam} optimizer with an initial learning rate of 0.001 and the batch size is set to 512. The whole training process lasts for a total of 500 epochs.

\subsection{Results}
\subsubsection{Emotion Recognition}

\paragraph{\textcolor{black}{(1) Binary Emotion Classification }} \textcolor{black}
{In our system, the neutral/non-neutral classification results from $T_{\mathcal{N}_{\mathit{eye}}}$ are serving as a trigger to capture emotional moments, and the accuracy of this binary classification can directly affect the performance of the whole system. Therefore, we first examine the quality of binary classification model EMO+ to ensure the trigger system is reliable enough. As shown in Table~\ref{tb::overrall_performance_binary}, the proposed EMO+ significantly outperforms the baseline EMO model and achieves \textcolor{black}{80.7\%} precision, \textcolor{black}{79.0\%} recall, and \textcolor{black}{80.4\%} accuracy on this binary classification task, 
This demonstrates the value of adding pupil information in EMO models. The high accuracy achieved by EMO+ also indicates that our \textit{EMOShip}-Net can be very sentiment to those emotional moments. }

\begin{table}[!htb]
\caption{\textcolor{black}{Performance comparison of binary emotion classification.}}
\label{tb::overrall_performance_binary}
\begin{tabular}{@{}cccccll@{}}
\toprule
Method &Precision & Recall & Accuracy   \\ \midrule
EMO+     &    \textcolor{black}{80.7\%}      &   \textcolor{black}{79.0\%}  &  \textcolor{black}{80.4\%}  \\ 
EMO   &     \textcolor{black}{78.1\%}       &   \textcolor{black}{74.6\%} &  \textcolor{black}{76.9\%}   \\ %
\bottomrule
\end{tabular}
\end{table}

\paragraph{\textcolor{black}{(2) Multiple Emotion Classification}} 
Table \ref{tb::overrall_performance} demonstrates the emotion recognition performance of the four baseline methods and our \textit{EMOShip}-Net on EMO-Film dataset. 

The performance of EMO \cite{wu2020emo} has significantly outperformed that of VinVL \cite{zhang2021vinvl} in terms of precision, recall, and accuracy. This is in line with our expectation, as the emotional clues within eye images are more generally straightforward when compared with the indirect and subtle sentimental clues in scene images. EMO+, the improvement version of EMO, has shown superior performance than EMO, and therefore the effectiveness of integrating pupil size information has been verified. The performance of VinVL+ also surpasses that of VinVL, which illustrates the importance of involving the user's attention. However, VinVL+ still cannot outperform EMO and EMO+, indicating the necessity of using expression-related features.

Different from those baselines, \textit{EMOShip}-Net fuses emotional evidence of both scene and eye images to achieve more comprehensive and accurate emotion predictions. Noticeably, EMOShip-Net has significantly outperformed the best baseline EMO+ by 5.3\% precision, 5.8\% recall, and 6.0\% accuracy. This reveals the necessity and importance of inferring from both expression and visual perceptions, and the superiority of \textit{EMOShip}-Net in terms of recognizing emotion states has been shown. 

\begin{table}[!htb]
\caption{Performance comparison \textcolor{black}{of multiple emotion classification} for the proposed method and the baseline methods.}
\label{tb::overrall_performance}
\begin{tabular}{@{}cccccll@{}}
\toprule
Method &Precision & Recall & Accuracy   \\ \midrule
\textit{EMOShip}-Net (Ours)     &    76.3\%      &  73.6\%  & 80.2\%  \\ 
EMO+   &    71.0\%       &  67.8\% & 74.2\%   \\ 
EMO \cite{wu2020emo}  &    65.9\%       &  67.0\% & 69.4\%   \\
VinVL+   &    48.8\%       &  46.8\% & 57.3\%   \\
VinVL \cite{zhang2021vinvl}   &    42.6\%       &  44.3\% & 55.5\%   \\

\bottomrule
\end{tabular}
\end{table}

\begin{figure}[h]
\begin{minipage}[t]{0.33\linewidth}
\subfloat[VinVL+]{\includegraphics[width=1 \textwidth]{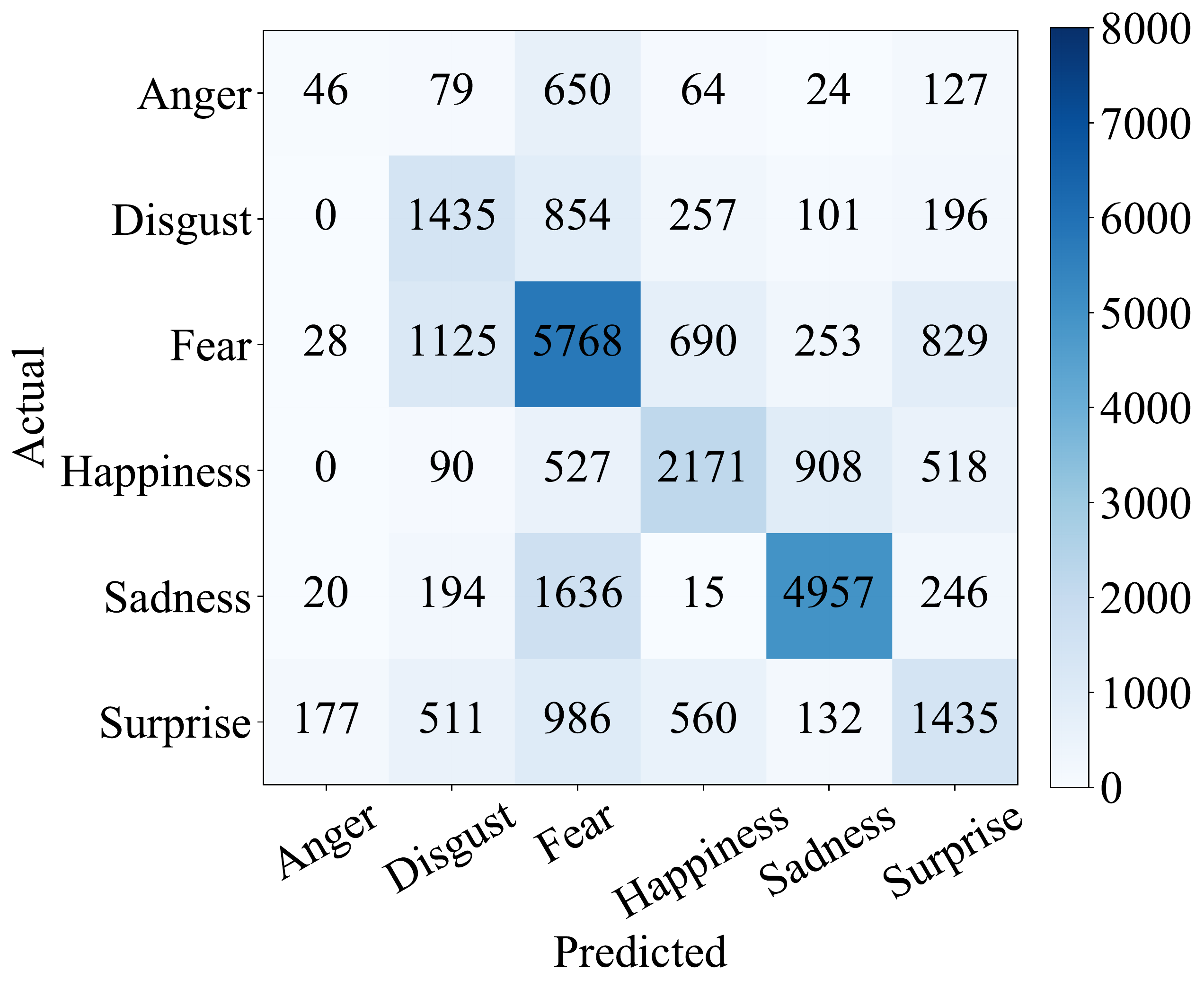}
\label{fig::cnfmtx_baseline1}}
\end{minipage}
\begin{minipage}[t]{0.33\linewidth}
\subfloat[EMO+]{\includegraphics[width=1 \textwidth]{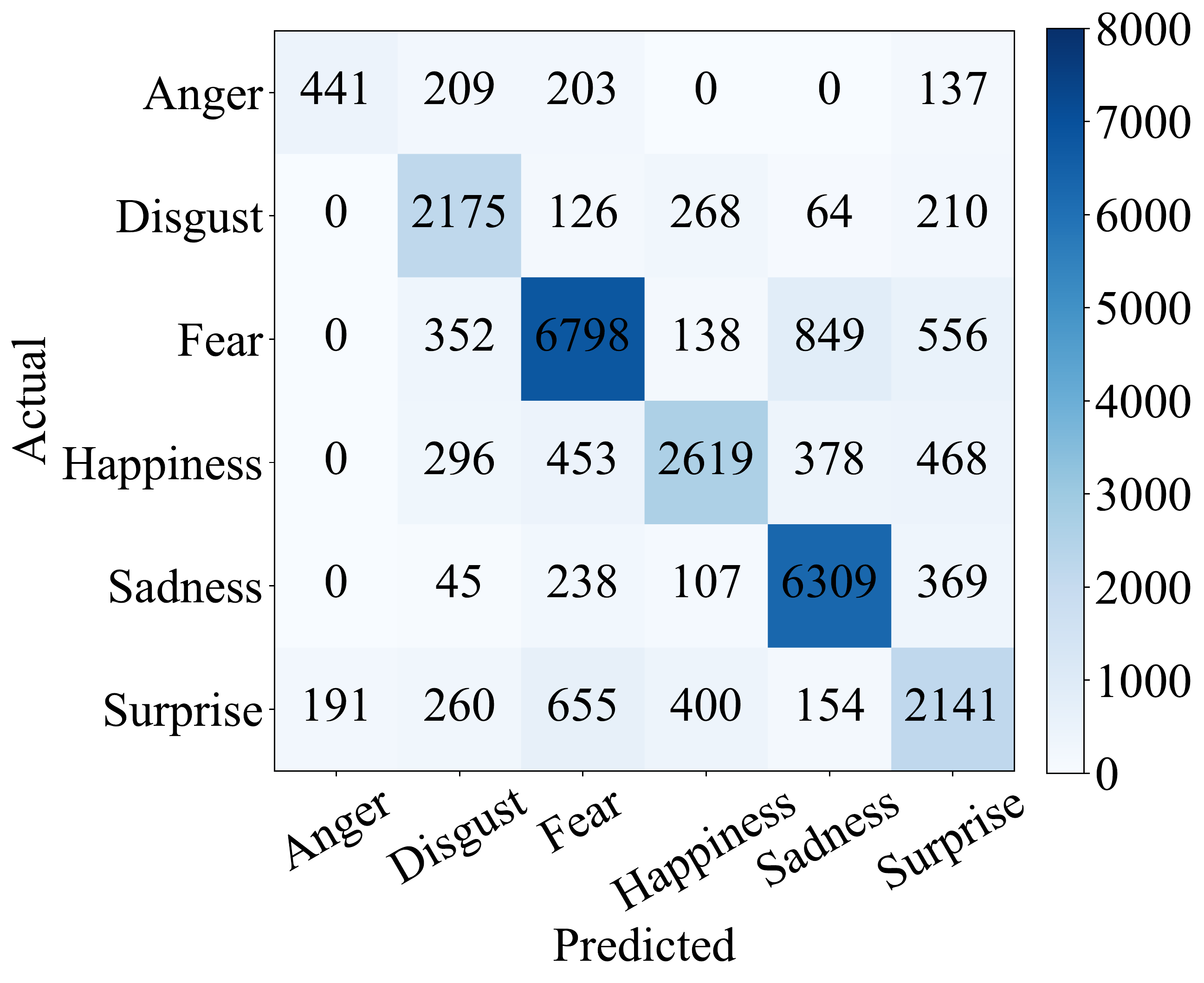}
\label{fig::cnfmtx_baseline2}}
\end{minipage}
\begin{minipage}[t]{0.33\linewidth}
\subfloat[\textit{EMOShip}-Net (Ours)]{\includegraphics[width=1 \textwidth]{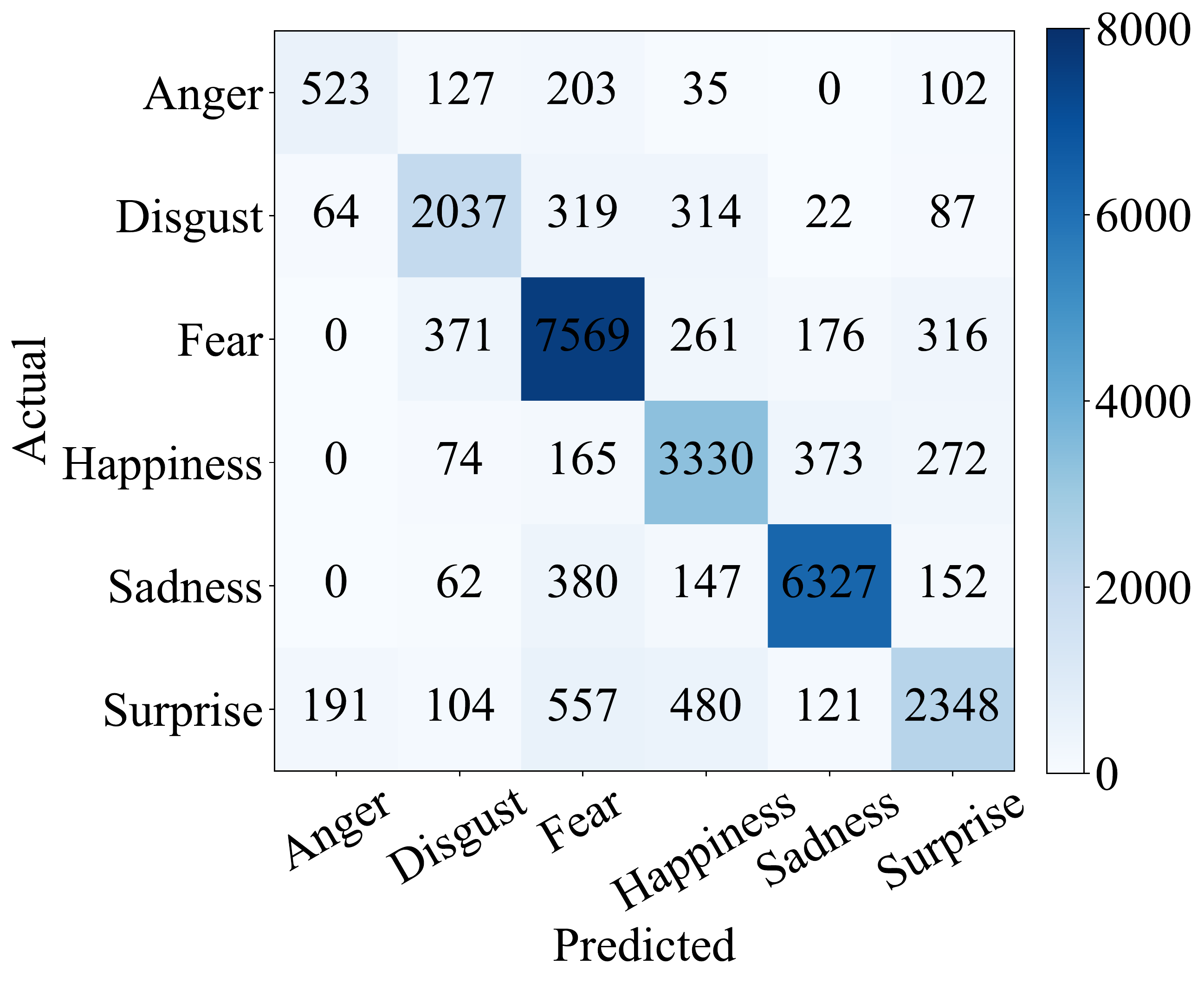}
\label{fig::cnfmtx_proposed}}
\end{minipage}
\caption{Confusion matrix of individual emotional moments when using the two baseline methods and the proposed method.}
\end{figure}

We have also plotted the confusion matrices of different methods to provide a more intuitive comparison. As shown in  Fig.~\ref{fig::cnfmtx_baseline1}, Fig.~\ref{fig::cnfmtx_baseline2}, and Fig.~\ref{fig::cnfmtx_proposed}, we discover that \textit{EMOShip}-Net achieves a better recognition rate on most of the emotions, demonstrating its superior generalization ability. \textcolor{black}{We can also observe that \textit{EMOShip} performs slightly worse on \say{disgust} than EMO+. That is because EMO+ determines emotional states exclusively based on eye images, while \textit{EMOShip} takes both the visual region and eye images into consideration. Therefore, \textit{EMOShip} may also suffer from this design when receiving strong misleading signals from visual attentive regions. A typical example is that when negatively-sentimental scene images are captured, it can be challenging for \textit{EMOShip} to determine which kind of negative emotions (such as \say{disgust} or\say{fear}) should be related to this visual information since they are all likely to happen as a result of viewing negative scenes. As shown in Fig.~\ref{fig::cnfmtx_baseline1} and Fig.~\ref{fig::cnfmtx_baseline2}, the VinVL+ method that only utilizes the visual information generally delivers lower classification rates on negative emotions such as \say{disgust} and \say{anger} than EMO+, while} their recognition accuracy on other classes, such as \say{happiness} and \say{sadness}, are relatively close with each other. \textcolor{black}{This validates that the associations between negative sentimental visions and negative emotions can be challenging to establish. }

\begin{figure}[h]
	\includegraphics[width=1 \textwidth]{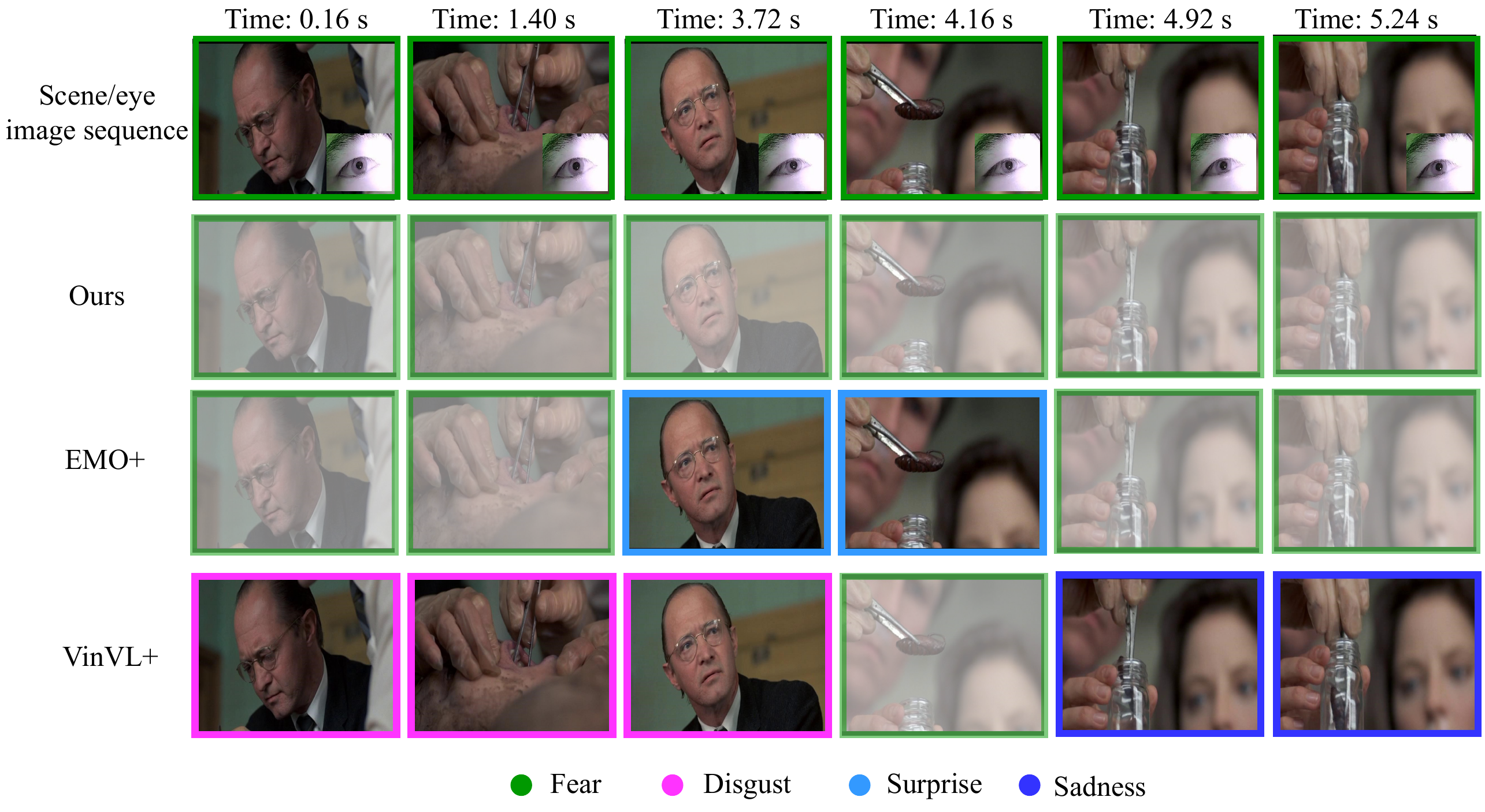}
 \caption{An example of emotion recognition comparison between the proposed \textit{EMOShip}-Net and the two baseline methods. We use colored rectangles to highlight the true emotions and the emotions predicted by these methods.}
 	\label{fig::case_emo_rcg}
\end{figure}

Fig. \ref{fig::case_emo_rcg} shows an exemplary case to provide further intuition. It presents
successive scene/eye image sequence and corresponding emotion predictions within an approximate six-second clip of \say{fear} emotions. 
Both VinVL+ and EMO+ baseline have produced inconsistent emotion predictions during this clip, while our method has successfully given the fear predictions for all those frames. This verifies that our method can also generate more temporal-consistent emotional predictions, thanks to the fusion of both shreds of emotional evidence from scene and eye regions.

\subsubsection{Understanding of Potential Cause of Emotions} 
Understanding the potential cause of emotions can be too subtle and subjective to be quantitatively evaluated, especially when we are aiming to compute the Influential Score $IS$, i.e. the degree of emotional impacts from visual perceptions. Despite those challenges, our \textit{EMOShip} is the first-ever eyewear system that is able to reveal the semantic attributes of visual attention and to associate those attributes with the varying emotional states. We analyze such behaviors through several representative examples. Particularly, we first adopt cases of different emotions to present the qualities of the summary tags of visual attentive regions, and then we plot the Influential Score from real video clips to examine their temporal patterns. 

\begin{figure}[h]
	\includegraphics[width=1.0 \textwidth]{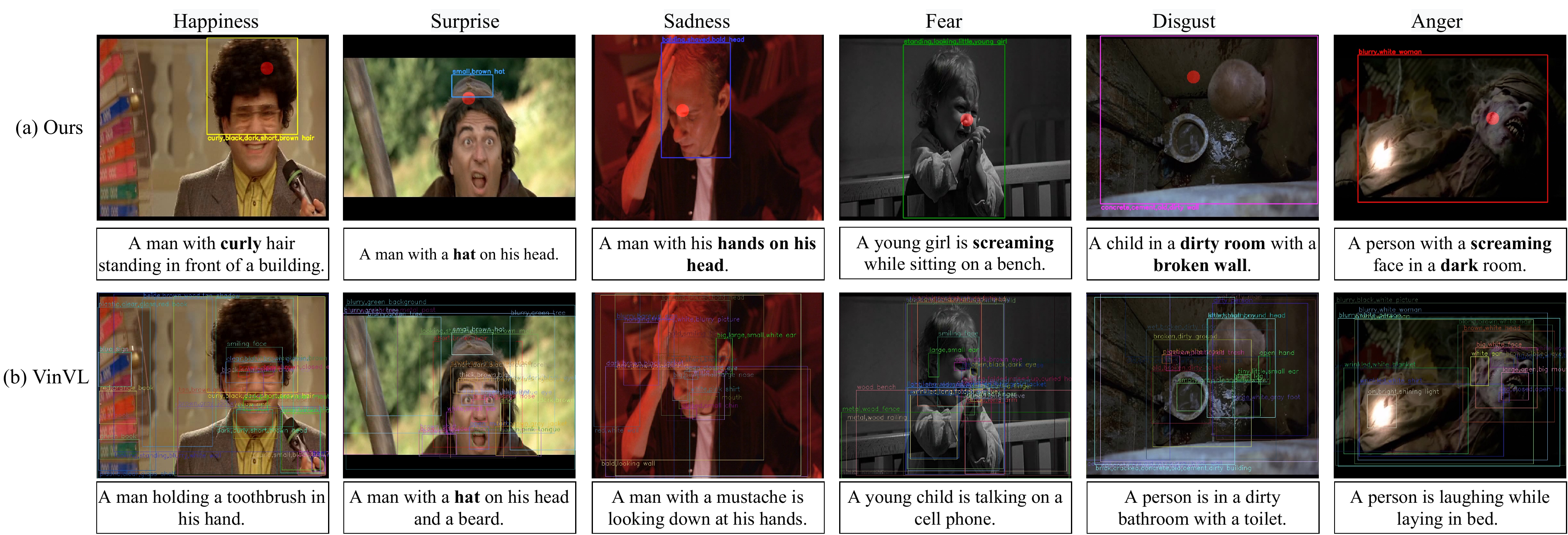}
	\caption{Examples of the semantic summary tags generated by our method (a) and the VinVL baseline (b). The summary tag is shown at the bottom of each frame. The red circle indicates the gaze point. The emotional words are highlighted in bold.} 
	\label{fig::nlp_qa}
\vspace{-1em}
\end{figure}

In Fig. \ref{fig::nlp_qa}, we provide examples of the semantic summary tags generated by our method and the VinVL baseline. 
A general trend can be discovered that the summary tag of our \textit{EMOShip} has better captured the sentimental clues in those scenarios than that of VinVL baseline.
For example, in the \say{fear} case, the summary tag of \textit{EMOShip} contains emotional-indicating keywords such as \say{screaming}, which is highly relevant with negative emotions like \say{fear} and is clear evidence of awareness of the sentimental visions. In contrast, VinVL displays neutral descriptions and use words like \say{talking} to depict this scene, which are less sentimentally accurate. Similar observations can be made on other emotions where our \textit{EMOShip} involved more emotional indicators such as \say{dirty room}, \say{screaming face}, \say{dark room}, etc.  \textcolor{black}{Those differences are not difficult to understand. The visual features used in the VinVL baseline method are not filtered and consist of visual information from non-attentive regions. Such irrelevant information can confuse the language model and can lead to less appropriate summary tags like the sentimentally neutral words. In contrast, \textit{EMOShip} 
utilizes the selected visual features (see Section \ref{sssec:select_roi_candidates} for details) that are highly relevant to the visual attentive region, and therefore the generated summary tags are more relevant with the visual attentive region and have higher chances to contain sentimentally non-neutral meanings.
}

Besides, the tags of our eyewear device are generally more semantically accurate than VinVL. Take \say{fear} for example. Our \textit{EMOShip} correctly depicts the scenario, i.e. \say{A young girl is screaming while sitting on a bench}, while VinVL has used some inappropriate words, that is, \say{A young girl is talking on a cellphone}. Such a kind of semantic accurateness is also an advantage of \textit{EMOShip}.

\begin{figure}[!htb]
	\centering
	\includegraphics[width=.5 \textwidth]{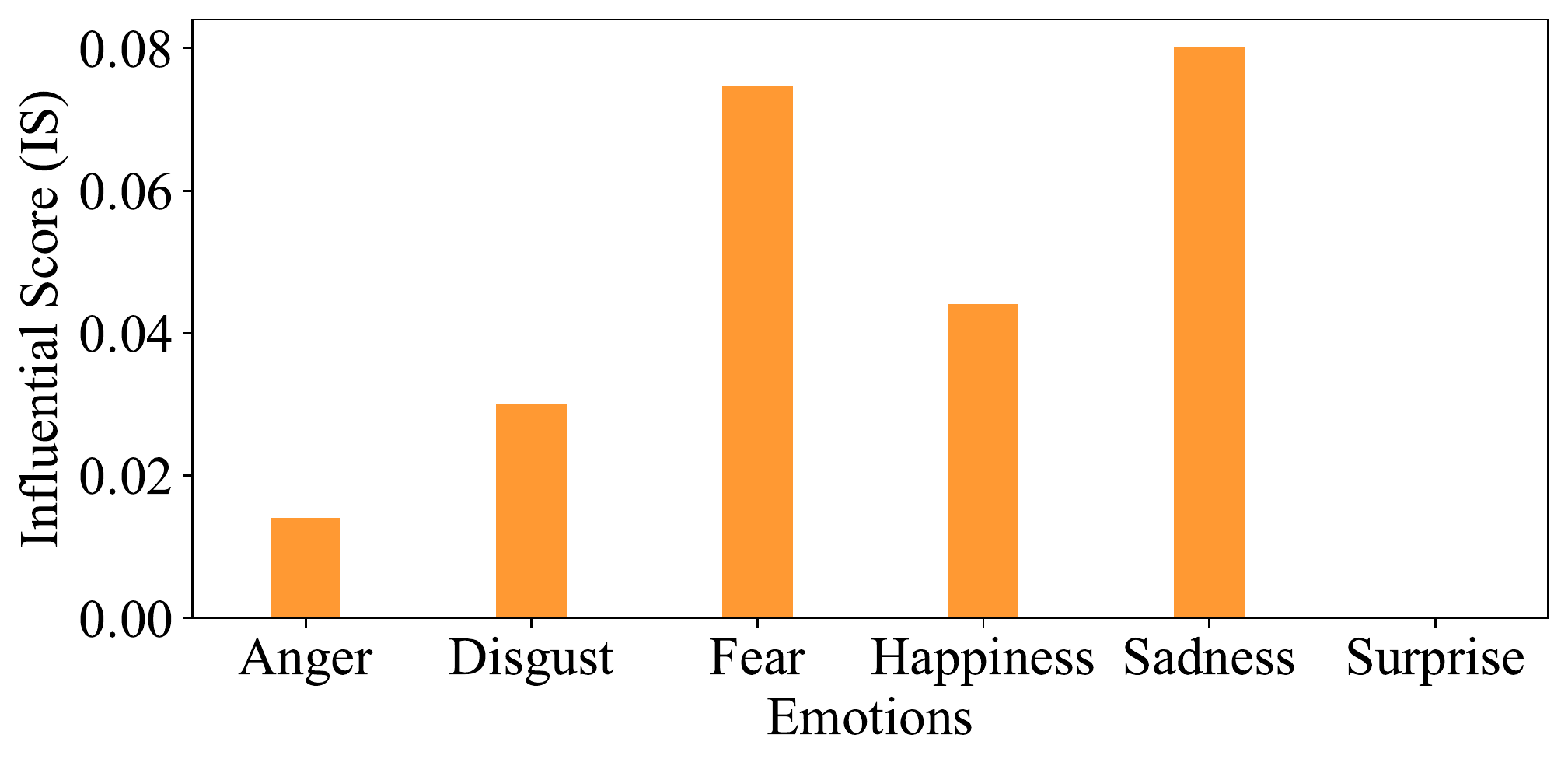}
	\caption[]{Degree of emotional impacts from visual perceptions.}
	\label{fig::IS-summary}
\end{figure}

\textcolor{black}{Next, we investigate how different emotional states can be associated with scene features through the use of Influential Score $IS$. Fig.~\ref{fig::IS-summary} shows the normalized average $IS$ of six non-neutral emotional categories. We can observe that the emotion \say{sadness} exhibits the highest the $IS$ value.
This indicates that emotion \say{sadness} is generally more tightly associated with our visual perceptions than others. Also, emotion \say{surprise} presents the lowest $IS$ score and is therefore considered to be less related with scene features than all other emotions.}

\subsubsection{\textcolor{black}{ Generalization Ability}}
\textcolor{black}{
We also examine the generalization ability of \textit{EMOShip}-Net on unseen users. 
Specifically, \textcolor{black}{5} new participants out of the EMO-Film dataset were recruited, and we follow identical data collection procedures as in Section \ref{sstn::data_collection} to formulate an extra evaluate set that is strictly subject-independent with the EMO-Film dataset of training our models. 
This new evaluation set totals up to \textcolor{black}{approximate 105} minutes, and then we evaluate the in-lab emotion recognition performance of \textit{EMOShip}-Net on this newly-collected unseen dataset. In particular, we compare the performance of \textit{EMOShip}-Net on this new test set with that of the EMO and EMO+ baseline methods (the two most out-standing baseline methods).
The results are shown in Table~\ref{tb::overrall_performance}. 
We can see that \textit{EMOShip}-Net demonstrates superior performance than EMO and EMO+. This exhibits that \textit{EMOShip}-Net can generalize well to unseen subjects. 
}
\begin{table}[!htb]
	\caption{\textcolor{black}{Performance comparison of multiple emotion classification for new/unseen users.} }
	\label{tb::overrall_performance}
	\begin{tabular}{@{}cccccll@{}}
		\toprule
		Method &Precision & Recall & Accuracy   \\ \midrule
		\textit{EMOShip}-Net (Ours) & \textcolor{black}{65.9\%}  & \textcolor{black}{76.6\%} & \textcolor{black}{78.5\%}  \\
		EMO+ & \textcolor{black}{62.8\%}  & \textcolor{black}{65.4\%} & \textcolor{black}{70.0\%}  \\ 
		EMO & \textcolor{black}{61.2\%}  & \textcolor{black}{60.3\%} & \textcolor{black}{61.4\%}  \\ 
		\bottomrule
	\end{tabular}
\end{table}

\textcolor{black}{We further examine the performance regarding F1 score of different methods on two test sets. Even for the same method, the performance on two different test sets is not very comparable, as one set may be more challenging than another. To validate the assumption, we compute F1 scores of different methods on the original/new test set along with the drop rates. As can be seen from Table~\ref{tb::f1}, all methods have shown a significant drop on this new test set, and therefore we may reasonably assume that this new test set can be a challenging set than the original one. Despite the raised difficulties, our proposed \textit{EMOShip}-Net still exhibits the lowest performance degradation on this new set, which verifies the importance of exploiting \textit{emotionship}.}
\begin{table}[!htb]
	\caption{\textcolor{black}{Performance comparisons regarding F1 score on subject dependent/independent test sets, respectively.}}
	\label{tb::f1}
\begin{tabular}{c|cc|c}
\hline
\multirow{2}{*}{Method} & \multicolumn{2}{c|}{F1 Score}                                                                                                                                                       & \multirow{2}{*}{Drop Rate} \\ \cline{2-3}
                        & \multicolumn{1}{c|}{\begin{tabular}[c]{@{}c@{}}Original Test Set\\ (subject-dependent)\end{tabular}} & \begin{tabular}[c]{@{}c@{}}New Test Set\\ (subject-independent)\end{tabular} &                            \\ \hline
EMOShip-Net (Ours)      & \multicolumn{1}{c|}{74.9\%}                                                                          & 70.8\%                                                                       & 5.4\%                      \\
EMO+                    & \multicolumn{1}{c|}{69.4\%}                                                                          & 64.1\%                                                                       & 7.6\%                      \\
EMO                     & \multicolumn{1}{c|}{66.4\%}                                                                          & 60.7\%                                                                       & 8.6\%                      \\ \hline
\end{tabular}
\end{table}

\section{Pilot Study}
\label{sctn::app}

In additional to in-lab experiments, we also perform \textcolor{black}{approximate three}-week in-field pilot studies to evaluate the performance of \textit{EMOShip} under realistic scenarios. 
In this section, we present two valuable real-life applications to fulfill the potentials of \textit{EMOShip} and to demonstrate its usability for everyday life, while we also discuss the current limitations of \textit{EMOShip} and the future works. 

\subsection{Applications}

The most significant advantage of \textit{EMOShip} is that it captures \textit{emotionship} instead of emotions. Compared with other emotional-aware glasses like EMO \cite{wu2020emo}, our \textit{EMOShip} not only predicts emotions at higher accuracy but also provides intuitive explanations on the potential causes of those emotional states. This awareness of \textit{emotionship} opens a gate into unseen attractive applications. 
Multiple rounds of user interviews lead to two representative and promising applications, which are  Emotionship Self-reflection and Emotionship Life-logging, respectively.

In psychology, the term \say{self-reflection} refers to the process of analyzing past behaviors to achieve better efficiency in the future \cite{ghosh2020towards,daudelin1996learning}. Self-reflection is indispensable, especially for people affected by negative emotions. As indicated in relevant studies \cite{di2018emotion}, negative emotions can lead to mental well-being issues. To maintain mental health, we need to self-reflect on negative emotional moments, and we also need to find what evokes those emotions such that our times of exposure to those causes can be intentionally minimized. 
This is a good scenario of applying \textit{EMOShip}. The ability to record emotional moments, to retrieve negative emotional moments, and to discover the emotional causes, can all be satisfied using \textit{EMOShip}. This application is named Emotionship Self-reflection.

Life-logging is usually considered as a digital self-tracking or recording of everyday life~\cite{selke2016lifelogging}, which is already a popular application. Current life-logging applications commonly record scenes with commercial glasses like GoPro and Google Clip, which lack personalized experiences. Besides, it is also difficult to categorize those recordings into different emotional categories, since those eyewear devices also lack emotion awareness. Manually classifying those emotional moments can be a practical way, yet it is extremely time-consuming and tedious, and the user may not be able to recall the extracted emotional activities. Therefore, we incorporated \textit{EMOShip} with life-logging to set up a new application Emotionship Life-logging. Different from conventional life-logging, our Emotionship Life-logging can automatically detect moments of different emotions and record those memorable moments as can be customized by wearers, and the potential emotional causes are also documented. Moreover, Emotionship Life-logging also enables various interesting and promising down-stream tasks such as emotional retrieving and classification \cite{yang2018retrieving}. 

\subsection{Procedure of Pilot Study} 
 In-field pilot studies are performed for those two applications. A total of \textcolor{black}{20} volunteers, including \textcolor{black}{14/6} males/females aged between 23 to 40, was recruited to participate in \textcolor{black}{pilot studies}. The research target of understanding the potential causes of emotions was briefed to all participants before the launching of the pilot studies. Those volunteers were also informed that their daily activities will be recorded for research purposes. 

During this in-field pilot study, participants were introduced to take on \textit{EMOShip} whenever they are convenient such that their everyday life can be covered as densely as possible. When equipped, \textit{EMOShip} automatically recorded those emotional moments along with the potential causes from visual attention. The complete scene videos taken by the world camera were also saved for reference and are referred to as the baseline video. 

The pilot studies lasted for around \textcolor{black}{three} weeks, and at the end of the study, volunteers were asked to assist the evaluation of \textit{EMOShip} in terms of understanding daily emotions and their causes. In particular, we require the participants: 1). to watch those emotional moments taken by \textit{EMOShip} and to mark those clips that they believed to have correctly reflected their emotional states, 2). also to retrieve from the baseline video those emotional moments that \textit{EMOShip} failed to capture. In addition, the participants were asked to complete a questionnaire survey regarding their opinions on the two emotionship applications, the usability of \textit{EMOShip}, the free feedback, etc.  

\begin{table}[!htb]
	\caption{Overall Performance of the in-field pilot study. \textcolor{black}{$T_{\mathit{always-on}}$ is the overall operation time of \textit{EMOShip}. $T_{\mathcal{N}_{\mathit{eye}}}$ and $T_{\mathit{capture}}$ are the operation time of eye tracking and the high-resolution video recording, respectively. EM means the emotional moments.}}
	\label{tb::acc_pilotstudy}
    \resizebox{\textwidth}{!}{
	\begin{tabular}{ccccccccccc}
		\hline
		Participant & \begin{tabular}[c]{@{}c@{}}$T_{\mathit{always-on}}$\\(minute)\end{tabular} &
		\begin{tabular}[c]{@{}c@{}}$T_{\mathcal{N}_{\mathit{eye}}}$ \\(minute)\end{tabular} &\begin{tabular}[c]{@{}c@{}}$T_{\mathit{capture}}$ \\ (minute)\end{tabular} &
		 \begin{tabular}[c]{@{}c@{}} \textcolor{black}{\# of Distinct}\\ \textcolor{black}{EM}\end{tabular} &
		 \begin{tabular}[c]{@{}c@{}} \# of True\\ EM\end{tabular} & \begin{tabular}[c]{@{}c@{}}\# of False\\ EM\end{tabular} & \begin{tabular}[c]{@{}c@{}}\# of Missed\\ EM\end{tabular} & Precision& Recall  \\ \hline
		P1          & 24.8 &  4.7    & 2.0  & \textcolor{black}{3}   & 17    & 3  & 2   &85.0\% & 89.5\%         \\
		P2          & 28.5 & 5.4    & 3.6  & \textcolor{black}{2}    & 10     & 2    &1  &83.3\% & 90.9\%  \\
		P3          & 42.5 & 2.7    & 2.0  & \textcolor{black}{1}     & 11  & 1 &2    &91.7\% &84.6\%     \\
		P4          & 55.6 & 7.0  & 2.1    & \textcolor{black}{3}   & 19   & 6  & 4   & 76.0\%& 82.6\%     \\ 
		P5          & 32.7 & 3.7  & 2.7     &\textcolor{black}{2}    &13    & 4  & 3  & 76.5\% & 81.3\%   \\
		P6          & 73.2 & 8.9    & 2.2   & \textcolor{black}{3}   & 23  &7   &4 & 76.7\% & 85.2\%  \\ 
		P7          & 44.7 & 2.4    & 1.2   & \textcolor{black}{1}  & 8    & 1  &0  &88.9\% &100.0\%    \\
		P8          & 17.9 & 1.9    & 1.1   &\textcolor{black}{1}   &8    	& 2  &3   &80.0\% & 72.7\%       \\ 
		P9          & 26.9 & 6.9     & 	1.2&\textcolor{black}{2}   & 9    & 3  & 3   &75.0\% &75.0\%\\ 
		P10         & 16.0 & 4.2     & 	1.5 &\textcolor{black}{1}   & 7    & 1 & 0 &87.5\% & 100.0\%   \\
	\textcolor{black}{P11} & \textcolor{black}{17.3} & \textcolor{black}{4.6}    & 	\textcolor{black}{1.4}& \textcolor{black}{4}   & \textcolor{black}{13}    & \textcolor{black}{2}  & \textcolor{black}{3}   &\textcolor{black}{86.7\%} &\textcolor{black}{81.3\%} \\ 
	\textcolor{black}{P12} & \textcolor{black}{10.9} & 
	\textcolor{black}{2.7}     & 	\textcolor{black}{0.6} &  \textcolor{black}{3}   &    \textcolor{black}{3} & \textcolor{black}{0}  &    \textcolor{black}{2}& \textcolor{black}{100.0\%} & \textcolor{black}{60.0\%} \\ 
	\textcolor{black}{P13} &\textcolor{black}{32.7} & \textcolor{black}{4.8}    & 	\textcolor{black}{2.0}& \textcolor{black}{4}   & \textcolor{black}{13}    & \textcolor{black}{1}  & \textcolor{black}{5}   &\textcolor{black}{92.9\%} &\textcolor{black}{72.2\%}\\ 
	\textcolor{black}{P14} & \textcolor{black}{14.5} & \textcolor{black}{2.4}    & 	\textcolor{black}{1.3}& \textcolor{black}{2}   & \textcolor{black}{9}    & \textcolor{black}{1}  & \textcolor{black}{0}   &\textcolor{black}{90.0\%} &\textcolor{black}{100.0\%}\\ 
	\textcolor{black}{P15} & \textcolor{black}{12.3} & \textcolor{black}{2.9}     & 	\textcolor{black}{1.5}& \textcolor{black}{3}   & \textcolor{black}{8}    & \textcolor{black}{1}  & \textcolor{black}{2}   &\textcolor{black}{88.9\%} &\textcolor{black}{80.0\%}\\ 
	\textcolor{black}{P16} & \textcolor{black}{14.2} & \textcolor{black}{3.2}     & 	\textcolor{black}{1.4}& \textcolor{black}{2}   &\textcolor{black}{8}    & \textcolor{black}{3}  & \textcolor{black}{1}   &\textcolor{black}{72.7\%} &\textcolor{black}{88.9\%}\\ 
	\textcolor{black}{P17} & \textcolor{black}{11.8} & \textcolor{black}{3.1}    & 	\textcolor{black}{1.2}& \textcolor{black}{1}   & \textcolor{black}{8}    & \textcolor{black}{0}  & \textcolor{black}{1}   &\textcolor{black}{100.0\%} &\textcolor{black}{88.9\%}\\ 
	\textcolor{black}{P18} & \textcolor{black}{24.2} & 
    \textcolor{black}{4.0}    & 	\textcolor{black}{2.2}& \textcolor{black}{2}   & \textcolor{black}{13}    & \textcolor{black}{2}  &\textcolor{black}{ 4}   &\textcolor{black}{86.7\%} &\textcolor{black}{76.5\%}\\ 
	\textcolor{black}{P19} & \textcolor{black}{14.3} & \textcolor{black}{3.2}     & 	\textcolor{black}{0.8}& \textcolor{black}{2}  & \textcolor{black}{4}    & \textcolor{black}{0}  & \textcolor{black}{2}   &\textcolor{black}{100.0\%} &\textcolor{black}{66.7\%}\\ 
	\textcolor{black}{P20} & \textcolor{black}{15.7} & \textcolor{black}{5.1}     & 	\textcolor{black}{1.8}& \textcolor{black}{3}   & \textcolor{black}{8}    & \textcolor{black}{4} & \textcolor{black}{1}   &\textcolor{black}{66.7\%} &\textcolor{black}{88.9\%}\\ 

		 \hline
		\textbf{Mean} &    &&  &	& & & & \textcolor{black}{\textbf{
			82.8\% }} &\textcolor{black}{\textbf{83.1\% }}  \\ \hline
			
\end{tabular}}
\end{table}

\subsection{Performance of \textit{EMOShip} in Pilot Study}

\subsubsection{Quantitative Evaluations}
Table~\ref{tb::acc_pilotstudy} summarizes the system performance of \textit{EMOShip}. Overall, the participants generated a total of \textcolor{black}{530.7} minutes of baseline video and \textcolor{black}{33.8} minutes of \textcolor{black}{212} video clips as their personal emotional moments. Compared with the overall operation time, \textcolor{black}{$T_{\mathit{always-on}} = 530.7 \mathit{min}$}, the operation time reduction for the eye features extraction and high-resolution video capturing are \textcolor{black}{84.2\% ($\frac{T_{\mathit{always-on}} - T_{\mathcal{N}_{\mathit{eye}}}}{T_{\mathit{always-on}} }$)
and 93.6\% ($\frac{T_{\mathit{always-on}} - T_{\mathit{capture}}}{T_{\mathit{always-on}} }$)}, respectively. \textcolor{black}{That is consistent with the short-term property of non-neutral emotions. As indicated in relevant research \cite{torkamaan2020mobile}, non-neutral emotions are typically aroused by a sudden emotional stimuli, and they are short-term mental processes that can vanish in a few seconds. In other words, non-neutral emotions are much rarer than neutral ones in our daily life. We inspect \textcolor{black}{P6} for a detailed understanding. \textcolor{black}{One of} the scenarios of \textcolor{black}{P6} is watching a \textcolor{black}{basketball game} lasting for around \textcolor{black}{12} minutes and our system has detected \textcolor{black}{0.4} minutes of non-neutral Emotional Moments (EM). 
Those EMs occurred exactly when the wearer has seen \textcolor{black}{two scoring shoots} that leads to his emotional reactions, each one lasting for around \textcolor{black}{0.2 minute.} Given a 30 fps sampling rate, the 2 EMs contain approximate 720 image frames (2$\times$30$\times$0.2$\times$60). Apart from those moments, \textcolor{black}{P6} remain emotionally neutral. \textit{EMOShip} correctly captures those non-netural emotional moments. }

Based on emotional moments marked by users at the end of the pilot study, we are able to evaluate \textcolor{black}{the performance of \textit{EMOShip} in practice. Generally, users pay attention to how many emotional moments are correctly recorded by \textit{EMOShip}, and how many emotional moments are missed or incorrectly recorded. We use precision $\frac{Number\; of \;True \;EM}{Number \;of\; True\; EM+ Number \; of \; False \; EM}$ to indicate the former, and recall $\frac{Number\;  of \; True \; EM}{Number \; of\;  True \; EM+ Number\;  of \; Missed\;  EM}$ to indicate the latter. Results show that, \textit{EMOShip} delivers \textcolor{black}{82.8\%} precision and \textcolor{black}{83.1\%} recall on average, which means that} \textit{EMOShip} can accurately capture the personal emotional moments, and most of the emotional moments can be captured by \textit{EMOShip}. 

\begin{figure}
	\centering
	\includegraphics[width=.5\textwidth]{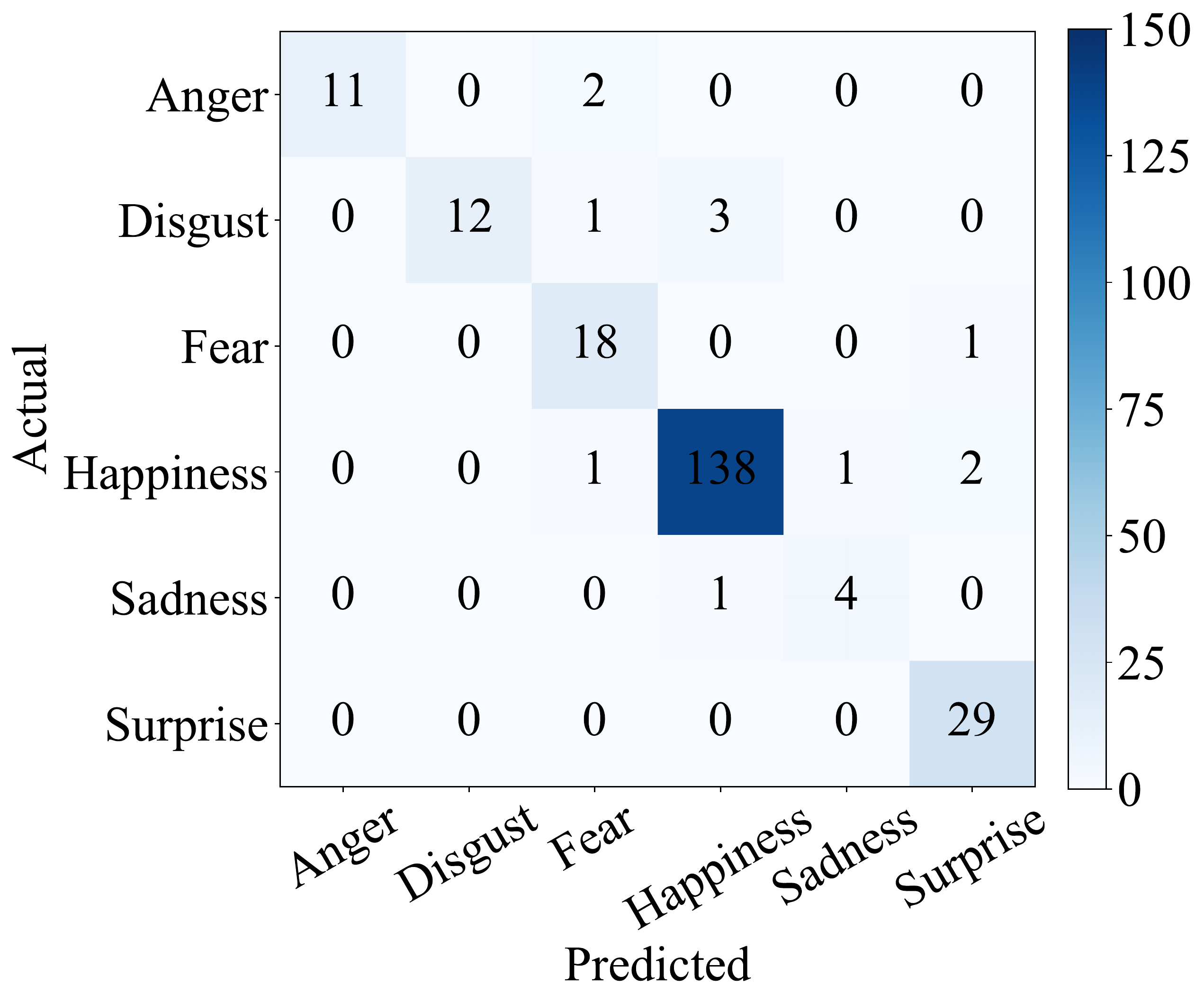}
	\caption{\textcolor{black}{Confusion matrix of individual emotional moments when using \textit{EMOShip} in pilot studies.}}	
	\label{fig::cnfmtx_pilot_study}
\end{figure}

\textcolor{black}{We also plot a confusion matrix on pilot studies to provide a more intuitive understanding. As shown in Fig.~\ref{fig::cnfmtx_pilot_study}, \textit{EMOShip} achieves a satisfying classification rate 
on the emotional categories.
Besides, positive emotions~\cite{christman1993equivalent} (\textcolor{black}{171} of \say{Happiness} and \say{Surprise}) are much more frequent than negative ones (\textcolor{black}{53} of \say{Sadness}, \say{Anger}, \say{Digust} and \say{Fear}), indicating that positive emotions are the dominating emotional states in daily life of those wearers during the pilot studies.}

\subsubsection{\textit{Emotionship} Analysis}
\begin{figure}[!htb]
	\includegraphics[width=1 \textwidth]{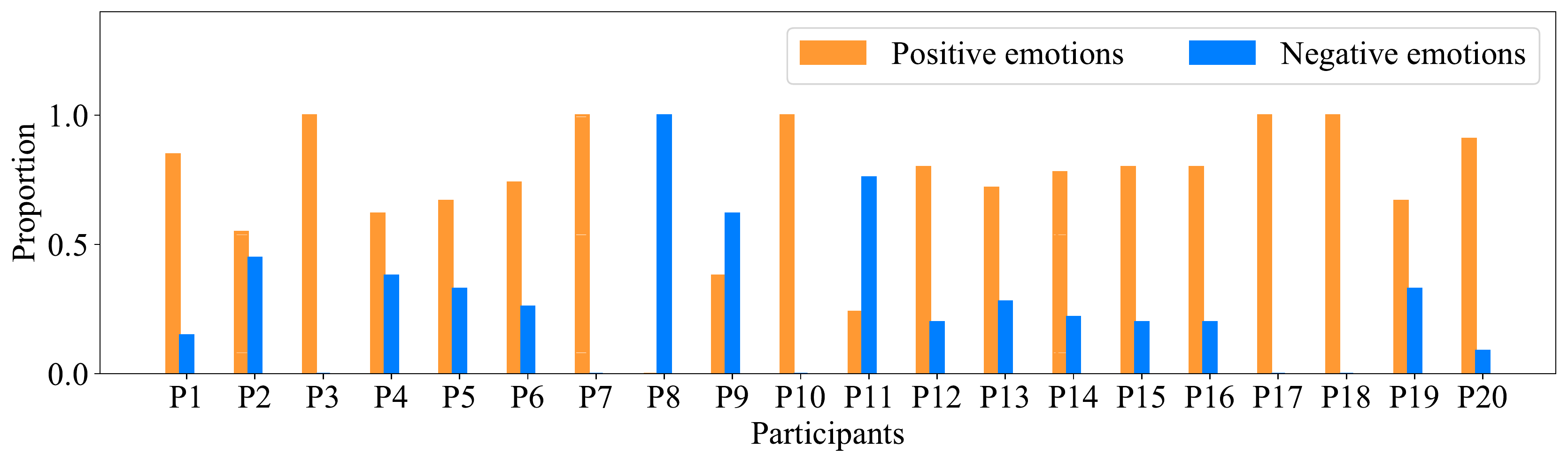}
	\caption{Profile of emotional states for all \textcolor{black}{20} participants.}
	\label{fig::emo_sum}
\vspace{-1em}
\end{figure}

\paragraph{(1) Emotional States Summary}
To give participants an overall understanding of their past emotional states, we briefly summarize the past emotional states for each user by roughly categorizing the six basic non-neutral emotional states as \textit{positive} and \textit{negative}. Intuitively, we categorize \say{happiness} and \say{surprise} as positive emotional states, while the rest four emotions are regarded as negative ones. 
For each user, we use $\mathit{Pr}$ and $\mathit{Nr}$ to denote the proportion of positive emotions and negative emotions, respectively. 
For a certain time window,
we can suggest two rough emotional patterns as follows:
\begin{itemize}
	\item Type \textrm{I}: $\mathit{Pr} >  \mathit{Nr}$, indicating that the overall emotional state of a user lean towards positive.
	\item Type \textrm{II}: $\mathit{Pr} \leq \mathit{Nr}$, suggesting that a user is more frequently occupied by negative emotions.
\end{itemize}
As shown in Fig.~\ref{fig::emo_sum}, we can observe that \textcolor{black}{17} out of \textcolor{black}{20} users belong to Type \textrm{I}, while \textcolor{black}{3} users fall into Type \textrm{II} (P8, P9, \textcolor{black}{and P11}), indicating that positive emotions are the dominating emotional states during the pilot studies. 

\begin{figure}[!t]
	\includegraphics[width=1 \textwidth]{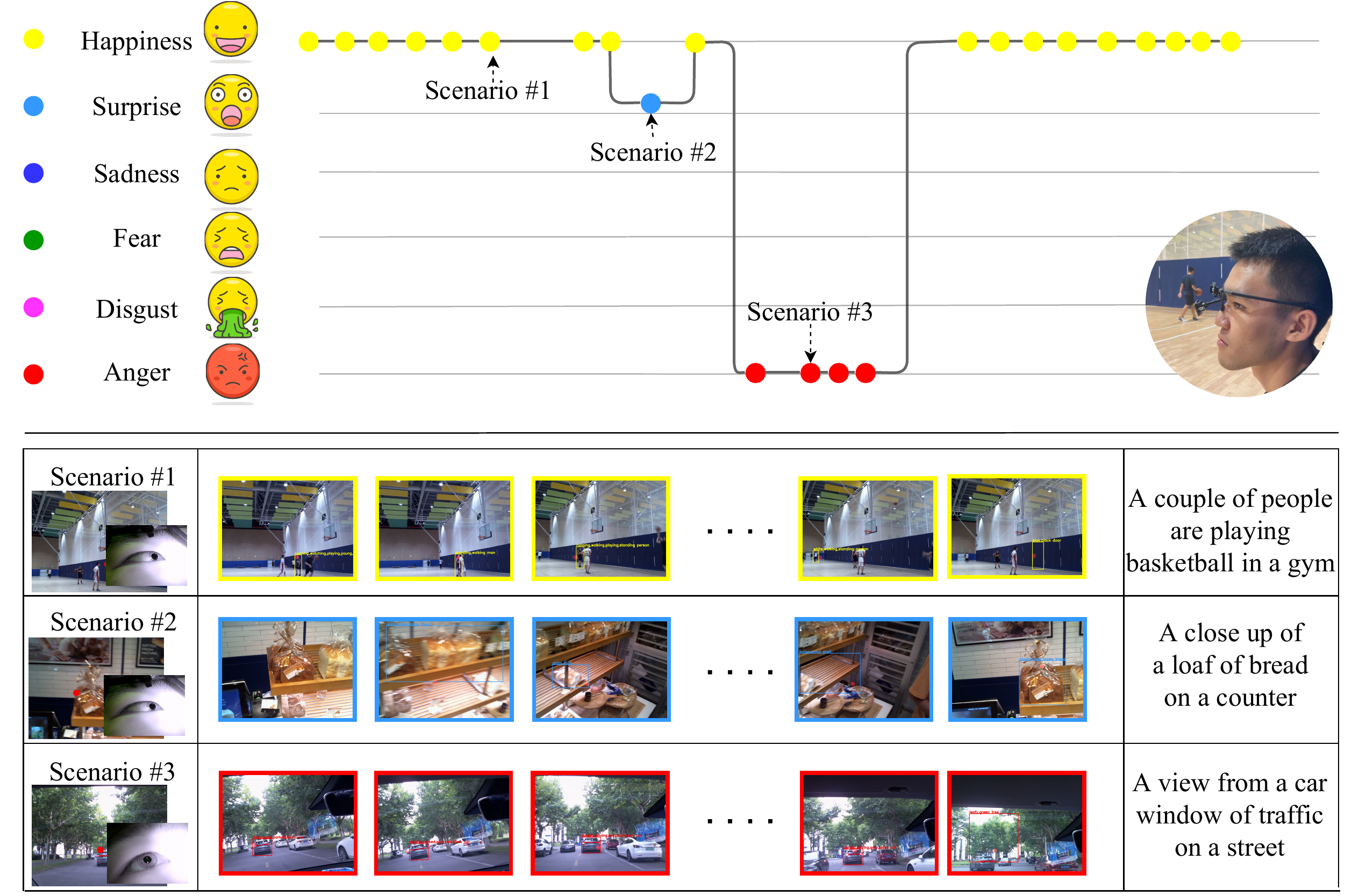}
	\caption{Time series emotional states for a participant (P6). Emojis are taken from~\cite{emoji2017}.}
	\label{fig::app1}
\vspace{-1em}
\end{figure}

\paragraph{(2) An Exemplary Case}
The following exemplary case provides further intuition on how \textit{EMOShip} works. 
Fig.~ \ref{fig::app1} shows the temporally-consistent emotional states for a participant (P6). 
The reason for selecting this subject is that P6 is the most active user during the pilot study, and P6's enthusiasm has led to 23 emotional clips with a duration time of 8.9 minutes covering most of the emotional categories, which provides us with good opportunities to explore the insight of \textit{EMOShip}.

As can be seen from Fig.~ \ref{fig::app1}, during the whole timeline, the major emotional state is \say{happiness}. This is not surprising, as we can examine the corresponding scenario, i.e. \say{Scenarios \#1} in the figure, and we can see from the summary tag that \say{A couple of people are playing the basketball in a gym}. This tag, along with the scenario image, indicates that this user is actually enjoying watching a basketball game and is quite likely to be happy. 
When it comes to the \say{surprise} case of \say{Scenarios \#2}, we can derive from the tag and also his attentive region that he is surprised to see close up of a loaf of bread.
As for the \say{anger} of \say{Scenarios \#3}, we can immediately learn from the summary tag and attentive region that he is driving a car and feels angry on traffic.
In a similar way, this participant can easily access all the emotional moments that are valuable and personalized. 
If this user would like to perform emotionship self-reflection, he may retrieve those anger moments, and discover what caused him to be anger, e.g. traffic, and he will intentionally avoid such scenarios, e.g. reduce the car driving frequency. In summary, \textit{emotionship} is a valuable asset and various promising applications are to be exploited.

\subsubsection{Feedback from Participants}
We also conduct a questionnaire survey to collect the comments from all participants on their opinions on the two applications, on their wearing experience, and on the potential improvement. 
In summary, \textcolor{black}{16 out of 20 } participants provided positive feedback on Emotionship Self-reflection, while \textcolor{black}{15} out of \textcolor{black}{20} people agreed on the values of Emotionship Life-logging, which are encouraging results. 

We randomly select the comments by several participants and quote them as follows.

One participant remarked:
\textit{``From my experience, EMOShip has allowed me to recognize and understand my emotions in a major meeting, which was quite profound to me. When I rewatched the video clips and emotions recorded by EMOShip, I realized that I appeared to be very negative during the meeting, and the meeting was also quite heavy. If I had noticed these issues then, I believe I would have been able to readjust myself to encourage the participation of the team and have a more productive meeting. So I think I will use EMOShip in more meetings and social events. In the long run, it would be significantly beneficial for me to understand and manage my emotions by utilizing EMOShip to analyze my emotions and record my emotional moments. ''---P1 }

Similarly, another participant appreciated the application of \textit{EMOShip} to long-term mood perception and management, as figured out by this volunteer:

\textit{``EMOShip shows that I have two significantly different states of mind when driving or walking to work. When I commute on foot, the emotions appear to be more positive and I tend to feel happy more frequently. My driving emotions, on the other hand, often seem to be negative, such as fear and anger. ...... I may feel negative or get road rage encountering rule-breaking behaviours such as slow left-lane driving or unsafe lane changes. In addition, with the help of EMOShip I also noticed that I seem to be overly cheerful during business meetings, which may leave an unintended impression of me being unprofessional or unreliable. EMOShip unveils the importance of facial expression management to me. I need to be more aware of my social environment whether I should be more happy or serious. ''---P2 }

The third user stated that \textit{EMOShip} can significantly ease the logging of emotional moments, 
which can be of importance:

\textit{``EMOShip can assist me to record some interesting or important moments and my emotions at that time, both of which are crucial for me to get these moments rapidly reviewed. ...... . Reviewing the meeting materials that are important to me by watching the videos EMOShip recorded can save me a great amount of time. Plus, my emotions may also shift during interesting moments in life. For example, EMOShip records intense game sessions and sensational videos when I feel happy or sad. It would have been very inconvenient for me to record them manually clip by clip while playing games or watching videos, whereas EMOShip can easily record them for me to review or share quickly afterwards. ''---P6}

On the other hand, there is also a volunteer who disregarded the importance of recording emotional moments, and we quote his feedback below: 
\textit{``I used EMOShip while playing cards. Since this is a highly enjoyable entertainment, there was little change in my recorded emotion types. Moreover, I probably didn't pay much attention to the changes in my emotions. ''---P7} 

\subsection{Limitations and Future Works}

We have demonstrated the technical capabilities of \textit{EMOShip} to recognize emotion states and understand their causes. However, we also observe several limitations from its applications to real-world scenarios and from users' feedback. In this section, we briefly discuss some potential future works that will further improve \textit{EMOShip} system.  

\subsubsection{Personalized Emotional Management}
Although most users have provided positive feedback on \textit{EMOShip}, a consensus is that they would like to also receive suggestions on how to reduce the occurrences of negative emotional moments. Since different people have different situations and hence requires personalized service, \textcolor{black}{we are planning to integrate into \textit{EMOShip} a long-term emotion tracking, emotional management, and regulation system which can be personalized to suggest how to avoid causes of negative emotions.}

\subsubsection{Privacy Concern and Privacy Protection}
Although participants are interested in perceiving their emotional states, some participants are uncomfortable with exposing their personal affective information to any third parties. Future system design should carefully consider how to address privacy concerns.
Another common feedback from users is that they are worried about the disclosing of their emotional information, especially to malicious third parties. Therefore, we set plans on enhancing the privacy protection of using \textit{EMOShip} and also on ensuring the safety of recorded personal data, from both software and hardware sides.

\subsubsection{Multi-Modality in Emotional Causes}
\textit{EMOShip} considers the visual stimuli as the most likely reason for stimulating emotions. However, visual perceptions are not the only perception that can arouse emotions. For example, the auditory perception, like a sharp, annoying sound, can also affect emotional states. How to fuse the emotional hints from multi-modality data remains a challenging topic for us to address in the future.

\section{Conclusions}
\label{sctn::cnclusn}
In this paper, we propose and address the \textit{emotionship} analysis problem for eyewear devices, an ambitious yet challenging task. To this end, we develop a deep network \textit{EMOShip}-Net that can predict the semantic attributes from scene images and can synthesis emotional clues from both eye and scene images with an awareness of each feature's importance factor. Based on \textit{EMOShip}-Net, we present \textit{EMOShip}, the first-ever intelligent eyewear system that is capable of \textit{emotionship} analysis. 
Experimental results on FilmStim dataset of 20 participants demonstrate that \textit{EMOShip} achieves 80.2\% emotional moment capturing accuracy, 
which significantly outperforms baseline methods.
We also demonstrate that \textit{EMOShip} can provide a valuable understanding of the cause of emotions.
We develop two promising applications using \textit{EMOShip}, i.e. emotionship self-reflection and emotionship life-logging, and we conduct \textcolor{black}{20} in-field pilot studies to demonstrate the usability and advantages.
Most participants like \textit{EMOShip} and think \textit{EMOShip} helps them realize and understand their emotions that they usually ignore. 
Most participants provide positive feedback on \textit{EMOShip} and admit its potentials. \textit{EMOShip} is the first practice of \textit{emotionship}-aware eyewear system, and it can be inspiring to the incoming age of intelligent wearable systems. Will smart glasses dream of sentiment visions? We will know soon.

\section{Acknowledgments}
\textcolor{black}{This work was supported in part by the National Natural Science Foundation of China under Grant No. 61932007 and 62090025 and in part by the National Science Foundation of the United States under grant CNS-2008151.}
\bibliographystyle{ACM-Reference-Format}
\bibliography{reference}
\end{document}